\documentclass[10pt,twocolumn,letterpaper]{article}

\usepackage[pagenumbers]{cvpr} 
\usepackage{array,multirow,graphicx}
\usepackage{makecell}
\usepackage{amssymb}
\usepackage{soul}
\usepackage{pifont}
\usepackage[accsupp]{axessibility}

\usepackage{float}
\newcommand{\STAB}[1]{\begin{tabular}{@{}c@{}}#1\end{tabular}}

\definecolor{cvprblue}{rgb}{0.21,0.49,0.74}
\definecolor{lightblue}{rgb}{0.9, 1, 1}
\usepackage[pagebackref,breaklinks,colorlinks,allcolors=cvprblue]{hyperref}
\usepackage{color, colortbl}

\def\paperID{15442} 
\def\confName{CVPR}
\def\confYear{2025}

\title{KeyFace: Expressive Audio-Driven Facial Animation for Long Sequences via KeyFrame Interpolation}

\author{Antoni Bigata$^{1}$ \qquad Michał Stypułkowski$^{2}$ \qquad Rodrigo Mira$^{1}$ \qquad Stella Bounareli \\
Konstantinos Vougioukas$^{1}$ \qquad Zoe Landgraf$^{1}$ \qquad Nikita Drobyshev$^{1}$ \\ 
Maciej Zieba$^{3}$ \qquad Stavros Petridis$^{1}$ \qquad Maja Pantic$^{1}$ \vspace{.2cm} \\
$^{1}$Imperial College London \qquad $^{2}$University of Wrocław \qquad $^{3}$Technical University of Wroclaw \\
{\tt\small ab4522@imperial.ac.uk}
}

\begin{document}
\maketitle
\begin{abstract}

Current audio-driven facial animation methods achieve impressive results for short videos but suffer from error accumulation and identity drift when extended to longer durations. Existing methods attempt to mitigate this through external spatial control, increasing long-term consistency but compromising the naturalness of motion. We propose \textbf{KeyFace}, a novel two-stage diffusion-based framework, to address these issues. In the first stage, keyframes are generated at a low frame rate, conditioned on audio input and an identity frame, to capture essential facial expressions and movements over extended periods of time. In the second stage, an interpolation model fills in the gaps between keyframes, ensuring smooth transitions and temporal coherence. To further enhance realism, we incorporate continuous emotion representations and handle a wide range of non-speech vocalizations (NSVs), such as laughter and sighs. We also introduce two evaluation metrics for assessing lip synchronization and NSV generation. Experimental results show that KeyFace outperforms state-of-the-art methods in generating natural, coherent facial animations over extended durations, successfully encompassing NSVs and continuous emotions. See the \href{https://antonibigata.github.io/KeyFace/}{project page} for visualizations and code.

\end{abstract}    
\section{Introduction}
\label{sec:intro}

\begin{figure}[ht]
  \centering
  \includegraphics[width=\linewidth]{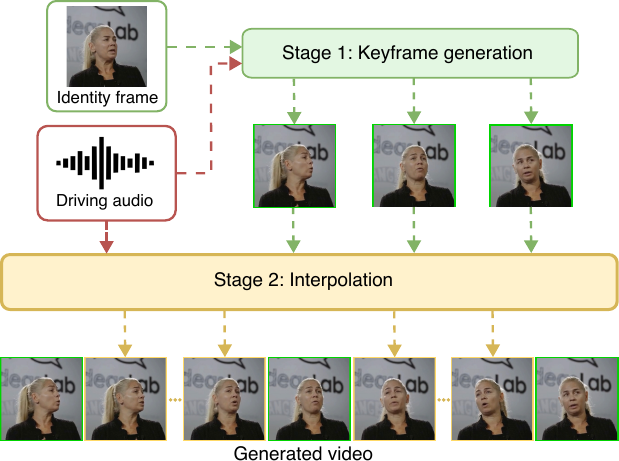}
  \caption{KeyFace generates long-term videos using a two-stage pipeline: first, keyframes are created as anchor points, then they are used by an interpolation model to produce smooth transitions.}
  \label{fig:intro_fig}
\end{figure}

The field of audio-driven facial animation has advanced significantly with the development of generative models like Generative Adversarial Networks (GANs)~\cite{goodfellow2020generative} and Diffusion Models (DMs)~\cite{ho2020ddpm, dhariwal2021diffusionmodelsbeatgans}. These approaches have greatly enhanced the realism and expressiveness of facial animations, enabling promising applications in virtual assistants, education, virtual reality, and aiding communication impairments~\cite{johnson2018assessing, kessler2018technology, wang2021one}. As a result, the demand for high-resolution, natural, long-term audio-driven facial animations has increased dramatically.

While early approaches in audio-driven facial animation were limited in terms of head rotation~\cite{Vougioukas} or focused solely on generating the mouth region~\cite{Prajwal}, current methods have advanced to produce results that are nearly indistinguishable from real videos. Despite this progress, most methods struggle when handling longer audio inputs, suffering from identity drift and overall quality degradation beyond the initial few seconds~\cite{xu2024hallohierarchicalaudiodrivenvisual, stypulkowski2023}. To extend generation length, some approaches incorporate additional spatial information, such as target head positions or landmarks, as model inputs~\cite{chen2024echomimiclifelikeaudiodrivenportrait, tian2024emoemoteportraitalive, wei2024aniportraitaudiodrivensynthesisphotorealistic}. While this can improve temporal consistency, it constrains animations to predefined facial motions, limiting expressiveness. Other methods use motion frames to provide context on prior movements~\cite{xu2024hallohierarchicalaudiodrivenvisual, stypulkowski2023}, but, as with many autoregressive approaches, small errors accumulate over time, reducing overall quality. 

Furthermore, recent methods often neglect important aspects of long-form natural speech, such as continuously changing emotions and NSVs. Existing emotional audio-driven methods often assume a fixed emotional state~\cite{ji2022eamm, tan2025edtalk, Gan_2023_ICCV}, which restricts them to short sequences and overlooks real-world dynamics, where emotions fluctuate continuously. Moreover, they typically rely on discrete emotion labels~\cite{gururani2023space, Gan_2023_ICCV}, which lack the nuance and fluidity of natural human expressions~\cite{wang2024improvement}. In contrast, the less-explored dimensions of valence and arousal provide a more precise portrayal of emotional states~\cite{sinem_aslan_2018, wang2024improvement}.
Similarly, NSVs such as laughter and sighs are largely neglected, despite being essential for natural communication~\cite{ruch2001expressive, pentland2010honest}. Crucially, handling emotion and NSVs requires a model capable of interpreting them over extended sequences to accurately animate the corresponding facial expressions.

To address these limitations, inspired by keyframe-based approaches~\cite{secondkey, thirdkey} initially introduced by~\cite{firstkey},  we propose \textit{KeyFace}, a novel two-stage approach for generating long and coherent audio-driven facial animations. In the first stage, a keyframe generation model produces a sequence at a low frame rate conditioned on an identity and audio input, spanning multiple seconds and eliminating the need for motion frames. In the second stage, an interpolation model fills in intermediate frames, ensuring smooth transitions and temporal coherence. By dividing generation into two parts, we implicitly separate motion and identity control, resulting in more natural motion and improved identity preservation over time. For longer sequences, this process can be repeated, with the interpolation model generating seamless transitions between segments. In addition to generating realistic, long-term animations, our pipeline allows for emotions that evolve over time, leveraging the keyframe generation model’s broad contextual span.

Our main contributions can be summarized as follows:

\begin{itemize}
    \item \textbf{State-of-the-art long-term animation}: We introduce a state-of-the-art method that combines keyframe generation with interpolation to produce videos that maintain high quality over time and capture long-range temporal dependencies.
    \item \textbf{Continuous emotion modelling}: Using valence and arousal, we enhance the emotional expressiveness of facial animations, allowing for nuanced portrayals of gradual emotional transitions.
    \item \textbf{Integration of non-speech vocalizations}: We extend the communicative capabilities of our model by incorporating NSVs for more natural animations.
    
\end{itemize}
\section{Related Works}
\label{sec:related}

\paragraph{Audio-driven facial animation}
Audio-driven facial animation methods~\cite{karras2017audio,Vougioukas,zhang2022sadtalker} generate realistic talking-head sequences with audio-synchronized lip movements. Early models, such as \cite{Vougioukas}, used GANs, introducing a temporal GAN to generate talking-head videos from a still image and audio input, while Wav2Lip~\cite{Prajwal} improved lip-sync accuracy with a pre-trained expert discriminator. More recent 3D-aware and head pose-driven methods~\cite{chen2020talking, zhang2022sadtalker, zhou2021pose} aimed to capture head motion, though often struggled with artefacts and unnatural movements.

In contrast to GANs, which face challenges like mode collapse~\cite{waster}, DMs excel in conditional image and video generation~\cite{rombach2022highresolutionimagesynthesislatent, zhang2023adding} and are promising for facial animation~\cite{du2023dae, xu2024vasa}. In \cite{stypulkowski2023}, an autoregressive diffusion model generate head motions and expressions from audio but face challenges with long-term consistency. Recent methods~\cite{xu2024hallohierarchicalaudiodrivenvisual, wang2024V-Express, chen2024echomimiclifelikeaudiodrivenportrait} use video DMs~\cite{ho2022video, blattmann2023stablevideodiffusionscaling} for improved temporal coherence. For instance, AniPortrait~\cite{wei2024aniportraitaudiodrivensynthesisphotorealistic} conditions on audio-predicted facial landmarks, but converting audio to latent motion (e.g., landmarks~\cite{wei2024aniportraitaudiodrivensynthesisphotorealistic} or 3D meshes~\cite{zhang2023dream}) remains challenging, often yielding synthetic-looking motion. Similarly, \cite{xu2024vasa} proposes a two-stage approach that disentangles motion and identity, but assumes strict separation, which is not always respected. To preserve identity across generated frames, several methods~\cite{wei2024aniportraitaudiodrivensynthesisphotorealistic, wang2024V-Express, chen2024echomimiclifelikeaudiodrivenportrait} leverage ReferenceNet~\cite{animateanyone}, which provides identity information, but increases resource demands. In contrast, KeyFace addresses these limitations by combining keyframe prediction and interpolation for temporally coherent, identity-preserving animations without relying on intermediate representations or ReferenceNet. Although similar approaches have been applied to video generation~\cite{make_a_video, blattmann2023stablevideodiffusionscaling} and controllable animation~\cite{ma2024followyouremojifinecontrollableexpressivefreestyle}, ours is the first to apply this method to audio-driven animation.

\paragraph{Emotion-driven generation}

Controllable emotion has recently become a key focus in audio-driven facial animation to create more realistic, empathetic avatars. Most works~\cite{gururani2023space, Gan_2023_ICCV, liang2022expressive, chen2023expressive} use discrete emotion labels (\eg. angry or sad) with intensity levels, but this approach lacks expressivity beyond predefined classes. Some approaches use a driving video or audio as a richer emotional source~\cite{tan2025edtalk, ji2022eamm, peng2023emotalk, lee2024speech, SaundersN23}, generating a latent representation from the media to drive the animation. This latent representation can sometimes be interpolated to control the resulting emotion~\cite{emovideo}. However, they require the driving audio or video during inference, limiting expressiveness and restricting explicit control. Continuous emotion conditioning, using valence and arousal, remains underexplored, despite evidence that it better captures emotional complexity~\cite{sinem_aslan_2018, wang2024improvement}. Additionally, few works allow for continuous emotion variation within a video, and those that do are often limited to a small set of emotions~\cite{multi_unified}, likely due to challenges in achieving coherent long-term animation.

\paragraph{Non-speech vocalizations}
Non-speech vocalizations (NSVs), such as laughter and sighs, significantly enhance human communication~\cite{pentland2010honest, ruch2001expressive} by providing context beyond words and increasing speech naturalness. Despite this, NSVs are often overlooked in audio-driven facial animation, and state-of-the-art models trained only on speech typically perform poorly on NSVs. Recently, two models have aimed to address this gap: Laughing Matters~\cite{casademunt2023laughing}, which proposes a diffusion model that can produce realistic laughter videos from still images and audio, and LaughTalk~\cite{sung2024laughtalk}, a 3D model that generates both speech and laughter. However, a model capable of handling multiple NSVs in addition to speech has not yet been explored.

\section{Method}
\label{sec:method}

\begin{figure*}
  \centering
  \includegraphics[width=\textwidth]{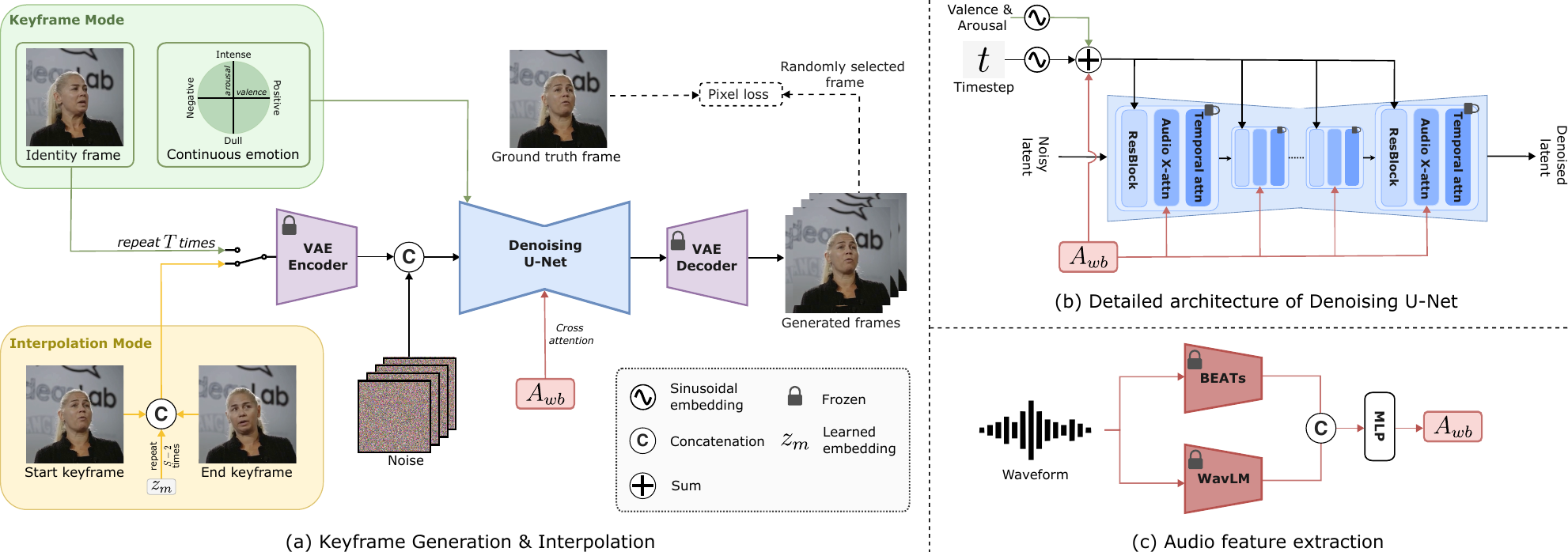}
    \caption{\textbf{Overview of KeyFace's two-stage framework.} The main architecture (a) is shared between the two stages, differing only in the conditioning inputs. A detailed view is provided in (b). In the keyframe generation stage, the model receives an identity frame $x_{\text{id}}$, repeated and concatenated with the noised video input to match the input dimensions. In the interpolation stage, the model is conditioned on two consecutive frames $z_{s}$ and $z_{e}$ from the keyframe sequence, interpolating the intermediate frames using a learned masked embedding $z_m$ and a binary mask $M$. Both stages incorporate audio embeddings $A_{wb}$ from WavLM and BEATs (c). We also use continuous emotion embeddings in the keyframe generation to produce facial animations that accurately convey both speech content and emotional expressions.}
    \label{fig:architecture}
\end{figure*}


Our two-stage approach, outlined in Fig.~\ref{fig:architecture}, starts with the generation of temporally distant keyframes. In the second stage, an interpolation model animates the full sequence by filling gaps between the generated keyframes. Our architecture builds upon Stable Video Diffusion (SVD)~\cite{blattmann2023stablevideodiffusionscaling}, with further architectural details and key distinctions provided in Appendix~\ref{supp:svd}.

\subsection{Latent diffusion} \label{sec:latent}
Diffusion models~\cite{ho2020ddpm, dhariwal2021diffusionmodelsbeatgans} are generative models structured as Markov chains with a Gaussian kernel, consisting of two main processes. The forward process gradually adds noise to the initial data point, while the reverse process denoises samples in multiple steps. Traditional diffusion models require many sampling steps to achieve high-quality images, which can be computationally demanding. The EDM framework~\cite{karras2022elucidatingdesignspacediffusionbased}, which defines the diffusion process as a stochastic differential equation and employs an Euler solver for denoising, reduces the necessary diffusion steps by parametrising the learnable denoiser $D_{\theta}$ as

\begin{equation}
    D_\theta(\mathbf{x}; \sigma) = c_{\text{skip}}(\sigma) \mathbf{x} + c_{\text{out}}(\sigma) F_\theta(c_{\text{in}}(\sigma) \mathbf{x}; c_{\text{noise}}(\sigma)),
\end{equation} where $F_{\theta}$ is the network to be trained, $\mathbf{x}$ is the model input, and $c_{\text{noise}}$, $c_{\text{out}}$, $c_{\text{skip}}$, and $c_{\text{in}}$ are scaling factors that depend on the noise level $\sigma$.
Latent Diffusion Models (LDMs)~\cite{rombach2022highresolutionimagesynthesislatent} further reduce computational demands by integrating a pre-trained Variational Autoencoder (VAE)~\cite{kingma2013auto}. Rather than operating in the original high-dimensional space, LDMs map data into a compact latent space via an encoder, where diffusion is applied more efficiently. The latent samples are subsequently decoded back to the original space.

\subsection{Keyframe generation}
\label{sec:keyframe}

In the first stage, we generate keyframes that capture essential facial expressions and movements, guided by the audio over an extended temporal context. These keyframes act as anchor points for the subsequent interpolation stage, ensuring that the final animation accurately reflects both the audio content and the associated emotional expressions. We generate $T$ keyframes, spaced $S$ frames apart, to capture long-range temporal dependencies efficiently.

Given a noised input sequence $z_k \in \mathbb{R}^{C\times T \times H \times W}$, where $C$ is the number of channels, and $H \times W$ are the spatial dimensions, our goal is to generate a sequence of a person speaking in sync with the given audio. To provide identity and background information, we repeat an identity frame $x_{id} \in \mathbb{R}^{C \times H \times W}$, pass it through the VAE encoder, and concatenate it with the noised input, effectively leveraging the U-Net architecture’s skip connections to preserve input details. Additionally, the model is conditioned on audio embeddings (see Section \ref{sec:audio}), along with emotional valence and arousal (see Section \ref{sec:emotion}).

\subsection{Interpolation}
\label{sec:interpolation}

After generating the main frames that capture essential facial expressions and movements, the next step is to interpolate between these keyframes to produce a smooth and coherent video sequence. 

We use the same architecture as the keyframe model, adapted for the interpolation task.
We take two consecutive frames $z_s$ and $z_e$ from the keyframe sequence as conditioning frames. To match the input shape $z_i \in \mathbb{R}^{C \times S \times H \times W}$, we create a sequence
\[
s = \{z_{s}, \underbrace{z_m, \dots, z_m}_{\text{repeat} \ S-2 \ \text{times}}, z_{e} \} \in \mathbb{R}^{C\times S\times H \times W},
\] where $z_m \in \mathbb{R}^{C\times H \times W}$ is a learned embedding that represents the missing frames. This sequence is concatenated channel-wise with the noise input. We also incorporate a binary mask $M \in \mathbb{R}^{S \times 1 \times H \times W}$, where $M_s = 1$ if $s$ corresponds to a conditioning frame ($s = 1$ or $s = T$), and $M_s = 0$ otherwise. This mask helps the model distinguish between conditioned and unconditioned frames, allowing it to focus on interpolating the intermediate frames.

\subsection{Audio encoding} \label{sec:audio}

For audio processing, we combine embeddings from two pre-trained audio encoders: WavLM $A_w \in \mathbb{R}^{L \times C^a}$~\cite{wavlm}, which excels at capturing linguistic content from speech, and BEATs $A_b \in \mathbb{R}^{L \times C^a}$~\cite{chen2022beatsaudiopretrainingacoustic}, which is trained to extract features from a broader range of acoustic signals, including non-speech sounds. We define $L \in \{T, S\}$ based on whether we use the interpolation or keyframe model and $C^a$ as the audio embedding dimension. By concatenating these embeddings, we obtain $A_{wb} = \text{Concat}(A_w, A_b) \in \mathbb{R}^{L \times 2C^a}$, which we feed to the model via two mechanisms:
\begin{itemize}
    \item \textbf{Audio Attention Blocks:} The combined embeddings serve as keys and values in the cross-attention layers within the U-Net, enabling the model to attend to relevant audio features.
    \item \textbf{Timestep Embeddings:} We pass $A_{wb}$ through an MLP and add it to the diffusion timestep embeddings $t_s \in \mathbb{R}^{C^s}$, yielding $t_s' = t_s + \text{MLP}(A_{wb})$, where $C^s$ is the timestep embedding dimension. This encourages alignment between image and audio frames.
\end{itemize}

\subsection{Emotion modelling with valence and arousal} \label{sec:emotion}

To capture complex, continuously changing emotional expressions, we adopt a continuous representation based on valence and arousal. For each frame, we extract valence and arousal using a pre-trained emotion recognition model~\cite{savchenko2022hsemotion}, encode them into sinusoidal embeddings $E_v, E_a \in \mathbb{R}^{C^s}$, and add them to the diffusion timestep embedding along with the audio embeddings: 

\begin{equation}
t_s'' = t_s' + E_v + E_a.
\end{equation}

Notably, we find that incorporating emotions solely in the keyframe model is sufficient for achieving effective emotional control, as the interpolation model can propagate emotional expressions without additional conditioning. During inference, users can provide any valence and arousal to guide the generation of desired emotional states.

\subsection{Losses}
\label{sec:losses}

Working in latent space is computationally efficient, but due to the compressed representations, it can be challenging for the model to retain fine semantic details from the original image~\cite{zhang2023show1marryingpixellatent}. This issue is particularly critical for faces, as humans are highly sensitive to minor imperfections in facial features, which can disrupt the perceived realism and emotional expressiveness of animations. To mitigate this, we decode the latent sequence $z_0$ back to RGB space to obtain $x_0 \in \mathbb{R}^{3\times L \times H \times W}$. We then apply an $L_2$ loss between the decoded frames $x_0$ and the ground truth frames $x_{\text{gt}}$, and add it to the existing $L_2$ loss between the latent representations $z_0$ and $z_{\text{gt}}$. We also include a perceptual loss $L_{p}$ based on features extracted from a pre-trained VGG network~\cite{johnson2016perceptuallossesrealtimestyle}, which encourages the generated images to be perceptually similar to the ground truth, enhancing visual quality.

To reduce memory consumption, we apply the additional pixel losses to a single random frame rather than the entire sequence, which proves sufficient for producing high-quality results. In contrast, the standard diffusion loss continues to be applied across all frames.  Moreover, we introduce a specialized weight $\lambda_{\text{lower}}$ applied to the lower half of the image, which helps the model focus on the mouth region. This spatially targeted weight, within the compressed latent space, enhances lip synchronization quality by emphasizing the alignment between generated lip movements and audio inputs, which is crucial for realistic audio-driven animations. The total loss function is defined as
\begin{equation}
    L = \lambda_{tot} \left(L_2(z_0, z_{\text{gt}}) + L_2(x_0, x_{\text{gt}}) + L_{p}(x_0, x_{\text{gt}}) \right),
\end{equation}
where $\lambda_{tot}=\lambda(t)\lambda_{lower}$ and $\lambda(t)$ is a weighting factor that depends on the diffusion timestep $t$, as defined in \cite{karras2022elucidatingdesignspacediffusionbased}.

\subsection{Guidance}
\label{sec:guidance}

For the keyframe model, we use a modified version of classifier-free guidance (CFG)~\cite{ho2022classifierfreediffusionguidance}, split into two parts: one for audio control and the other for identity control, allowing separate scales for each. The guidance formula is
\begin{equation} \label{eq:D_function}
        z = z_{\emptyset} + w_{\text{id}} \cdot (z_{\text{id}} - z_{\emptyset}) + w_{\text{aud}} \cdot (z_{\text{id \& aud}} - z_{\text{id}}),
\end{equation}
where $w_{\text{aud}}$ and $w_{\text{id}}$ are the guidance scales for audio and identity, respectively. Here, $z_{\emptyset}$ is the model output with all conditions set to 0, $z_{\text{id}}$ is the output with only the identity condition, and $z_{\text{id \& aud}}$ is the output with both conditions.

While CFG is effective in many scenarios, it can overly amplify the conditioning signal, reducing output diversity~\cite{karras2024guidingdiffusionmodelbad}. In the interpolation stage, where subtle emotional expressions and fluid motion are critical, CFG may be ill-suited. Autoguidance~\cite{karras2024guidingdiffusionmodelbad}, addresses this by using a model that is either smaller or trained with fewer steps to guide the main diffusion process, balancing guidance for improved video quality without sacrificing diversity. Autoguidance is formulated as
\begin{equation}
    D(x; \sigma, c) = D_{\text{r}}(.) + w_{\text{auto}} \cdot \left( D_{\text{m}}(.) - D_{\text{r}}(.) \right),
\end{equation}
where $D(x; \sigma, c)$ is the guided denoising step, $D_{\text{r}}(.)$ is the reduced model, $D_{\text{m}}(.)$ is the fully trained guiding model, and $w_{\text{auto}}$ controls the guiding model’s influence.

\section{Experiments}
\label{sec:experiments}

\subsection{Datasets}
\label{sec:data}

We train both models on HDTF~\cite{hdtf} and a dataset that we collected, comprising 160 hours of speech and 30 hours of NSVs. We also experiment with CelebV-Text~\cite{yu2022celebvtext} and CelebV-HQ~\cite{zhu2022celebvhq} but find that excluding these lower-quality datasets benefits training. For testing, we use the HDTF test set, and 100 videos randomly selected from CelebV-Text. Additionally, to evaluate our emotion control, we use a test set selected from MEAD~\cite{kaisiyuan2020mead}, as in \cite{tan2025edtalk}.

\subsection{Evaluation metrics}
\label{sec:metrics}

We evaluate image quality using the aesthetic quality metric from VBench~\cite{huang2023vbenchcomprehensivebenchmarksuite}, Fréchet Inception Distance (FID)~\cite{DBLP:conf/nips/HeuselRUNH17}, and Learned Perceptual Image Patch Similarity (LPIPS)~\cite{zhang2018unreasonableeffectivenessdeepfeatures}. For general video quality, we use Fréchet Video Distance (FVD)~\cite{unterthiner2019accurategenerativemodelsvideo} and the smoothness metric from \cite{huang2023vbenchcomprehensivebenchmarksuite}.
We compute the emotion accuracy ($Emo_{acc}$) using the pre-trained emotion recognition model from \cite{savchenko2022hsemotion}. We also introduce two new metrics, further details are provided in Appendix~\ref{supp:metrics}.

\paragraph{LipScore.} The typical metric for audio-visual synchronization, SyncNet~\cite{Prajwal}, has known limitations, including low correlation with lip-sync quality and significant reliability issues even on ground truth data~\cite{drobyshev2024emoportraitsemotionenhancedmultimodaloneshot, gururani2023space, yaman2024audiovisualspeechrepresentationexpert}. To address this, we introduce a lipreading perceptual score (LipScore) inspired by~\cite{lele2020}, which computes the cosine similarity between the generated and ground truth embeddings extracted from the final layer of a state-of-the-art lipreader~\cite{Ma_2023}. This lipreader is trained on 6$\times$ more data than SyncNet, providing higher-quality embeddings that better correlate with human perception.

\paragraph{NSV accuracy.} To evaluate the model’s ability to generate NSVs, we train a video classifier to recognize 8 NSV types ("Mhm", "Oh", "Ah", coughs, sighs, yawns, throat clears, and laughter) plus speech, for a total of 9 classes. The classifier is based on a pre-trained MViTv2~\cite{li2022mvitv2improvedmultiscalevision} for video classification on the Kinetics dataset~\cite{kay2017kineticshumanactionvideo}. We then employ it to evaluate the model's ability to generate the correct NSV type and measure the overall accuracy, denoting this metric as NSV accuracy  ($NSV_{acc}$).

\subsection{User study}

To provide a more comprehensive evaluation, we conduct a user study inspired by~\cite{chiang2024chatbot}. We select 20 videos per model (see Table~\ref{tab:metrics_comparison}) and present participants with pairs of 5-second videos featuring the same audio and identity, asking them to choose the more realistic video based on visual quality, lip synchronization, and motion realism.  We surveyed 51 participants, each of whom compared an average of 20 video pairs. We use an Elo rating system~\cite{elo1978rating} with bootstrapping applied to obtain a more stable ranking~\cite{chiang2024chatbot}.

\section{Results}

This section presents a comprehensive evaluation of our model, including comparisons with established methods and ablation studies to assess the impact of key components.

\subsection{Quantitative analysis}

\begin{table*}[ht]
\centering
\begin{tabular}{llccccccc}
\toprule
                                          & \multirow{2}{*}{Method}      & \multirow{2}{*}{\makecell{AQ $\uparrow$}} & \multirow{2}{*}{\makecell{FID $\downarrow$}} & \multirow{2}{*}{\makecell{LPIPS $\downarrow$}} & \multirow{2}{*}{\makecell{FVD $\downarrow$}} & \multirow{2}{*}{\makecell{Smoothness $\uparrow$}} & \multirow{2}{*}{\makecell{LipScore $\uparrow$}} & \multirow{2}{*}{\makecell{Elo $\uparrow$}} \\ \\ \midrule
\multicolumn{1}{c}{\multirow{6}{*}{\STAB{\rotatebox[origin=c]{90}{HDTF}}}} & SadTalker~\cite{zhang2022sadtalker}   & 0.52   & 60.55   & 0.44 &   410.86    &  \textbf{0.9955}   &  0.24    &   960.44              \\
\multicolumn{1}{c}{}                      & Hallo~\cite{xu2024hallohierarchicalaudiodrivenvisual}         & 0.55   &    \underline{19.22}   &   \underline{0.17}    &   236.97  &  0.9939    & 0.27   &    \underline{1054.69}         \\
\multicolumn{1}{c}{}                      & V-Express~\cite{wang2024V-Express}    & 0.55   & 34.68    &  0.21   &  \underline{200.67}   &  0.9943      &   \textbf{0.37}      &   985.35     \\
\multicolumn{1}{c}{}                      & AniPortrait~\cite{wei2024aniportraitaudiodrivensynthesisphotorealistic}   & \underline{0.56}   &  20.68   &  0.19    &  299.09   &  0.9951      &  0.14  & 887.84      \\
\multicolumn{1}{c}{}                      & EchoMimic~\cite{chen2024echomimiclifelikeaudiodrivenportrait}    &  0.55  &  20.35   &   0.18    &   213.30  & 0.9928       & 0.17   &  1023.53     \\
\multicolumn{1}{c}{}                      & \cellcolor{cvprblue!40}KeyFace &  \cellcolor{cvprblue!40}\textbf{0.59}   &  \cellcolor{cvprblue!40}\textbf{16.76}   &  \cellcolor{cvprblue!40}\textbf{0.16}    & \cellcolor{cvprblue!40}\textbf{137.25}   &  \cellcolor{cvprblue!40}\underline{0.9952}      & \cellcolor{cvprblue!40}\underline{0.36}   &    \cellcolor{cvprblue!40}\textbf{1091.52}     \\ \hline
\multirow{6}{*}{\STAB{\rotatebox[origin=c]{90}{CelebV-Text}}}             & SadTalker~\cite{zhang2022sadtalker}   &  0.49       &   49.85    &  0.49   &   434.31     &  \textbf{0.9959}  &   0.25 & 950.56  \\
                                          & Hallo~\cite{xu2024hallohierarchicalaudiodrivenvisual}     &  0.50  & 24.86    &  0.27     &  310.00   &  0.9938      &  0.29  &  1020.27   \\
                                          & V-Express~\cite{wang2024V-Express}     &  0.51  &  26.46   &  \underline{0.22}     &  \underline{253.16}   &   0.9933     &  \textbf{0.32}  &  \underline{1044.43}    \\
                                          & AniPortrait~\cite{wei2024aniportraitaudiodrivensynthesisphotorealistic}    & \underline{0.52}   &  24.84 & 0.28  &     373.32  &  0.9950   &    0.12    &       841.79     \\
                                          & EchoMimic~\cite{chen2024echomimiclifelikeaudiodrivenportrait}   &  0.51  &     \underline{22.81} &  0.26     &    298.33 &  0.9921      &  0.18  &  1043.26    \\ 
                                          & \cellcolor{cvprblue!40}KeyFace & \cellcolor{cvprblue!40}\textbf{0.55}   & \cellcolor{cvprblue!40}\textbf{17.06}   & \cellcolor{cvprblue!40}\textbf{0.21}      &  \cellcolor{cvprblue!40}\textbf{180.26}   &   \cellcolor{cvprblue!40}\underline{0.9952}     &  \cellcolor{cvprblue!40}\underline{0.30}  &  \cellcolor{cvprblue!40}\textbf{1100.90}   \\
                                         \bottomrule
\end{tabular}

\caption{\textbf{Quantitative comparisons} on HDTF~\cite{hdtf} and CelebV-Text~\cite{yu2022celebvtext} between our model and state-of-the-art facial animation methods. The best results are highlighted in \textbf{bold}, and the second-best results are \underline{underlined}. All the metrics are described in Section~\ref{sec:metrics}}
\label{tab:metrics_comparison}
\end{table*}

\begin{figure}[ht]
  \centering
  \includegraphics[width=\linewidth]{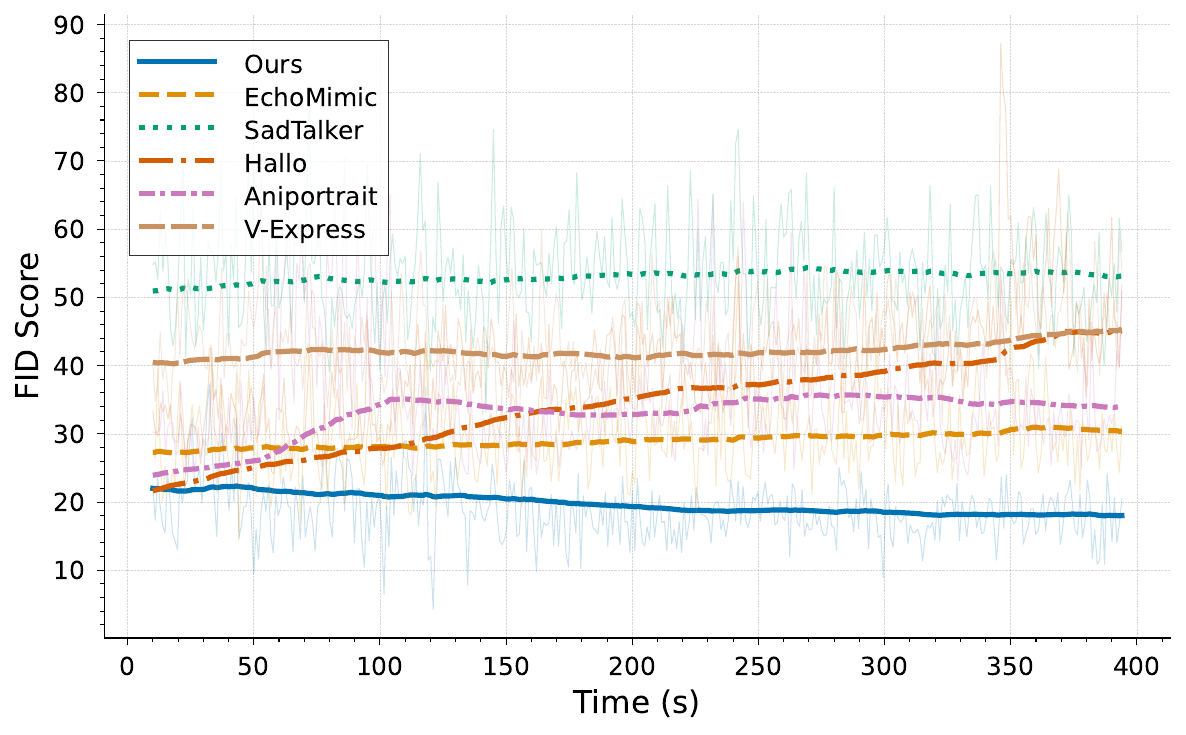}
  \caption{We present sliding window FID with a 1-second window size for videos generated by different methods.}
  \label{fig:fid_over_time}
  \end{figure}

We present a quantitative comparison against current state-of-the-art methods in Table~\ref{tab:metrics_comparison}. KeyFace achieves the lowest FID and FVD, indicating higher realism and temporal coherence. Our model also achieves the highest AQ and LPIPS, confirming the visual appeal of our animations. While SadTalker and V-Express achieve the highest smoothness and LipScores, respectively, KeyFace ranks a close second and outperforms both SadTalker and V-Express on the other metrics, demonstrating better performance overall. Figure~\ref{fig:fid_over_time} illustrates FID over time for videos generated by each method, where our two-stage approach maintains consistent quality without degradation. In contrast, Hallo and AniPortrait suffer from significant quality loss over time. To ensure fairness in evaluation, we also report results for a variant of our model trained exclusively on HDTF in Appendix~\ref{supp:only_hdtf}.

Additionally, the user study results (Elo) show that our model is preferred over other methods, confirming the effectiveness of our approach. A detailed analysis of the results is provided in Appendix~\ref{supp:user}.

\subsection{Emotional results}

To assess our model’s ability to generate accurate emotional expressions, we evaluate it on MEAD~\cite{kaisiyuan2020mead}, comparing it to state-of-the-art models in Table~\ref{tab:emo_comparison}.

\begin{figure}[ht]
  \centering
  \includegraphics[width=\linewidth]{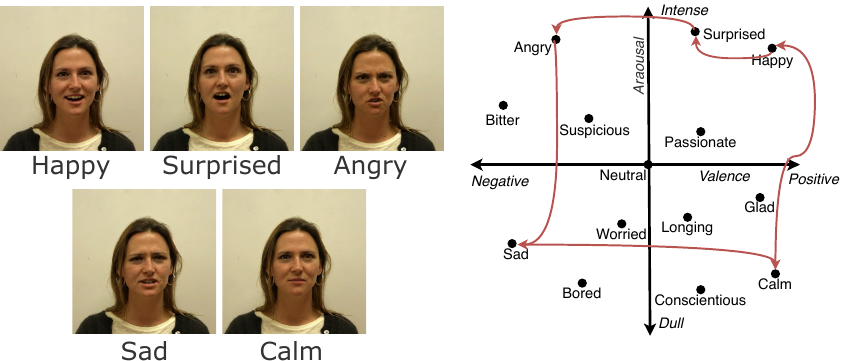}
  \caption{We show KeyFace's ability to interpolate between several different emotions within the same video.}
  \label{fig:emo_interp}
\end{figure}

Notably, despite being trained with only pseudo-labels extracted from our training data, KeyFace achieves competitive emotion accuracy compared to other models, which are trained on ground-truth labels from MEAD, outperforming 2 out of 3 models while delivering significantly better image and video quality. We also show that using continuous emotion labels (valence and arousal) yields significant improvements compared to discrete labels, and allows our model to generate multiple emotions within a single video by interpolating between points in the valence and arousal space, as illustrated in Figure~\ref{fig:emo_interp}.

\begin{table}[ht]
\resizebox{\columnwidth}{!}{%
\begin{tabular}{lcccc}
\toprule
                          Method    & Emotion Source & FID $\downarrow$ & FVD $\downarrow$ & $Emo_{acc}$ $\uparrow$ \\ \midrule
EDTalk~\cite{tan2025edtalk}             &    Video        &   101.19  &  619.90   &   \textbf{0.72}       \\
EAT~\cite{Gan_2023_ICCV}  & Discrete Labels  &  75.69   &  560.61   &   0.54       \\
EAMM~\cite{ji2022eamm}      & Video               &  107.16   &  855.20   &   0.17       \\
KeyFace &  Discrete Labels   &  \underline{50.34}   &  \underline{509.13}   &  0.43        \\
\rowcolor{Gray!40} KeyFace & Valence \& arousal &  \textbf{44.43}   & \textbf{447.74}    &    \underline{0.67}      \\ \bottomrule
\end{tabular}}
\caption{\textbf{Emotion evaluation} on MEAD~\cite{kaisiyuan2020mead}. Default settings are highlighted in \colorbox{Gray!40}{gray} on all tables.}
\label{tab:emo_comparison}
\end{table}

\subsection{Ablation Studies} \label{sec:ablations}


\paragraph{Audio Encoder.} We evaluate the impact of different audio encoders on the model’s ability to handle both speech and non-speech vocalizations (NSVs). As shown in Table~\ref{tab:abl_audio}, the combination of WavLM and BEATs achieves the best overall performance. BEATs is shown to significantly improve the handling of NSV, aligning with the findings of~\cite{casademunt2023laughing}. Likewise, when it is removed, the ability to generate the correct NSVs becomes close to random probability. Finally, WavLM improves lip synchronization quality compared to Wav2Vec2, with only a slight sacrifice in image quality, as indicated by a marginal increase in FID.

\begin{table}[ht]
\centering
\resizebox{\columnwidth}{!}{%
\begin{tabular}{lcccc}
\toprule
Audio backbone             & FID $\downarrow$& FVD $\downarrow$& LipScore $\uparrow$ & $NSV_{acc}$ $\uparrow$ \\ \midrule
WavLM       &  16.89  &  147.12   &  \textbf{0.36} & 0.10  \\
BEATs          &  19.52  &  212.47   &  0.29 & 0.23  \\
Wav2vec2 + BEATs       &  \textbf{16.00}   &   \underline{143.97}  &  \underline{0.32} & \underline{0.31}  \\ 
\rowcolor{Gray!40} WavLM + BEATs          &  \underline{16.76}   &   \textbf{137.25}  &  \textbf{0.36} & \textbf{0.42}  \\
\bottomrule
\end{tabular}}
\caption{\textbf{Audio encoder ablation} on HDTF~\cite{hdtf}. For $NSV_{acc}$, we use HDTF identities with audio containing NSVs.}
\label{tab:abl_audio}
\end{table}

\paragraph{Architecture.} We assess two key architectural modifications in Table~\ref{tab:temporal_ablation}. First, we examine the effect of removing temporal layers for keyframe generation, which leads to overly static frames, highlighting the importance of generating keyframes as a cohesive sequence. Second, we replace the concatenation operation with ReferenceNet, inspired by recent trends from~\cite{animateanyone}, and find that it requires twice as many training steps to achieve acceptable results. Even then, it results in lower video quality, introducing inconsistencies in background continuity and face shape.

\begin{table}[ht]
\centering
\resizebox{\columnwidth}{!}{%
\begin{tabular}{lccc}
\toprule
Method             & FID $\downarrow$& FVD $\downarrow$& LipScore $\uparrow$ \\ \midrule
w/o temporal layers &  \underline{23.74}   &  \underline{250.30}   &  0.25    \\
w/o Concat, w/ Reference Net       &  39.71   &   401.70  &  \underline{0.32}    \\ 
\rowcolor{Gray!40} Concat w/ temporal layers     &  \textbf{16.76}   &   \textbf{137.25}  &  \textbf{0.36}   \\ 
\bottomrule
\end{tabular}}
\caption{\textbf{Architecture ablation} on HDTF~\cite{hdtf}.}
\label{tab:temporal_ablation}
\end{table}

\paragraph{Losses.} We compare the effects of different pixel loss functions and weights in Table~\ref{tab:loss_ablation}. First, we see that having a pixel-space loss proves to be beneficial regardless of the loss type. The L1 loss yields improved image quality and lip synchronisation, but restricts model flexibility compared to the L2 loss, resulting in a higher FVD. In addition, incorporating the $L_{p}$ loss noticeably improves visual quality for both losses, as shown by the decreased FID. Overall, combining $L_2$ and $L_{p}$ losses produces the best balance of quality, variety, and lip synchronisation. Next, we examine $\lambda_{lower}$, which controls the weight of the pixel loss for the lower part of the video. Choosing a higher weight improves animation quality and lip synchronization, but increasing it too much overemphasizes this region and reduces overall quality. A good balance is achieved with $\lambda_{lower}=3$.

\begin{table}[ht]
\centering
\begin{tabular}{lccc}
\toprule
Method             & FID $\downarrow$& FVD $\downarrow$& LipScore $\uparrow$ \\ \midrule
No pixel loss       &  18.76   & 148.22    &  0.33   \\
$L_1$ only          &  17.66   &   172.01  &  \textbf{0.37}   \\
$L_2$ only          &  19.00   &  \underline{137.54}   & 0.34    \\
$L_1$ + $L_{p}$  &  \underline{17.02}   &  169.01   &  0.34    \\ 
\rowcolor{Gray!40} $L_2$ + $L_{p}$  &  \textbf{16.76}   &   \textbf{137.25}  &  \underline{0.36}   \\ 
\midrule
$\lambda_{lower}=1$        &  17.40   &  186.87   & 0.33   \\
$\lambda_{lower}=2$        &   \textbf{16.71}  &  \underline{147.01}   &  \underline{0.35}  \\
\rowcolor{Gray!40} $\lambda_{lower}=3$  &  \underline{16.76}   &  \textbf{137.25}   &  \textbf{0.36}   \\
$\lambda_{lower}=4$        &   17.36  &  161.40   &  \underline{0.35}  \\
\bottomrule
\end{tabular}
\caption{\textbf{Loss ablation} on HDTF~\cite{hdtf}.}
\label{tab:loss_ablation}
\end{table}

\paragraph{Data.} We test the effect of adding additional training data to each stage in Table~\ref{tab:data_abl}. Both models experience a decline in performance when trained with data of lower quality, confirming our hypothesis from Section~\ref{sec:data}. We find that training exclusively on high-quality data primarily improves lip synchronization for the interpolation model, while conversely enhancing video quality for the keyframe model, as indicated by lower FID and FVD scores, respectively. This suggests that each model plays a distinct role in the generation process and therefore reacts differently to changes in training data.

\begin{table}[ht]
\resizebox{\columnwidth}{!}{%
\centering
\begin{tabular}{ccccc}
\toprule
          \multicolumn{2}{c}{Training set}  & \multirow{2}{*}{FID $\downarrow$} & \multirow{2}{*}{FVD $\downarrow$} & \multirow{2}{*}{LipScore $\uparrow$}    \\
         \cmidrule(lr){1-2}
            Keyframe        & Interpolation       &&& \\ \midrule
                                  All   &     All      &  26.92   &  253.24   &  0.24   \\
                                  All   &     HQ only      &  24.45  &   236.75  & \underline{0.31}    \\
                      HQ only         &        All   &  \underline{16.97}   &  \underline{166.81}    &  0.24   \\ 
          \rowcolor{Gray!40} HQ only           &      HQ only    &   \textbf{16.76}  &  \textbf{137.25}   & \textbf{0.36}   \\
\bottomrule
\end{tabular}
}
\caption{\textbf{Data Ablation} on HDTF~\cite{hdtf}. “HQ only” refers to our high quality training set (HDTF and collected data), while “All” refers to all training data, including CelebV-Text and CelebV-HQ.}
\label{tab:data_abl}
\end{table}

\paragraph{Guidance.} 

\begin{figure}[ht]
  \centering
  \includegraphics[width=\linewidth]{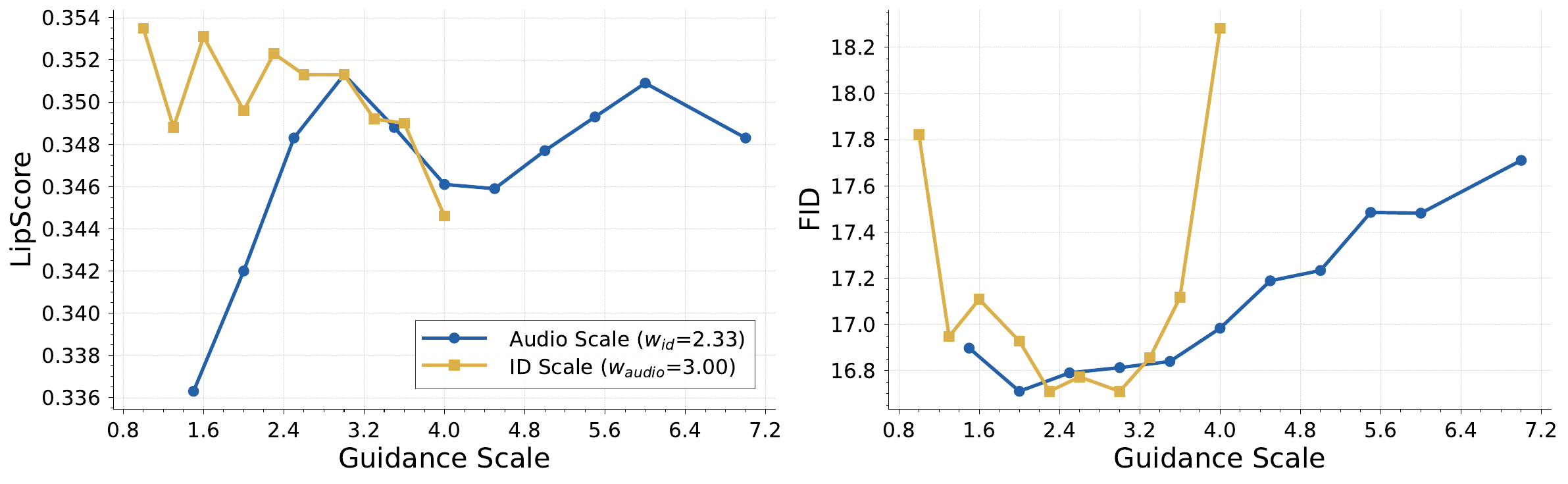}
  \caption{We show the impact of \textbf{guidance scale} for identity and audio condition on FID and LipScore on HDTF~\cite{hdtf}.}
  \label{fig:guidance_weight}
\end{figure}

We compare different guidance types in Table~\ref{tab:inf_abl}. Using CFG for both models makes videos overly static, as it closely adheres to the keyframes, limiting expression range and animation flow, as shown by the lower FVD. In contrast, applying autoguidance to the keyframe model worsens alignment with audio, resulting in lower LipScores. Using CFG instead allows for a separate grid searches for audio and identity guidance scales (Fig.~\ref{fig:guidance_weight}), increasing flexibility and enhancing model performance.

\begin{table}[ht]
\resizebox{\columnwidth}{!}{%
\begin{tabular}{ccccc}
\toprule
\multicolumn{2}{c}{Guidance method}   & \multirow{2}{*}{FID $\downarrow$} & \multirow{2}{*}{FVD $\downarrow$} & \multirow{2}{*}{LipScore $\uparrow$}    \\
\cmidrule(lr){1-2}
Keyframe     & Interpolation &&& \\ \midrule
  Autoguidance~\cite{karras2024guidingdiffusionmodelbad}   & CFG~\cite{ho2022classifierfreediffusionguidance} & 20.12   &  172.31   & 0.31 \\
Autoguidance~\cite{karras2024guidingdiffusionmodelbad} & Autoguidance~\cite{karras2024guidingdiffusionmodelbad}       &  18.86   & \underline{152.77}    & \underline{0.33}   \\
CFG~\cite{ho2022classifierfreediffusionguidance}       & CFG~\cite{ho2022classifierfreediffusionguidance}         &  \underline{18.53}   &  177.09   &  0.32  \\ 
\rowcolor{Gray!40} CFG~\cite{ho2022classifierfreediffusionguidance}     & Autoguidance~\cite{karras2024guidingdiffusionmodelbad} &  \textbf{16.76}   & \textbf{137.25}    &  \textbf{0.36}  \\
\bottomrule
\end{tabular}}
\caption{\textbf{Guidance Ablation} on HDTF~\cite{hdtf}.}
\label{tab:inf_abl}
\end{table}

\subsection{Qualitative analysis}

\begin{figure}[ht]
  \centering
  \includegraphics[width=\linewidth]{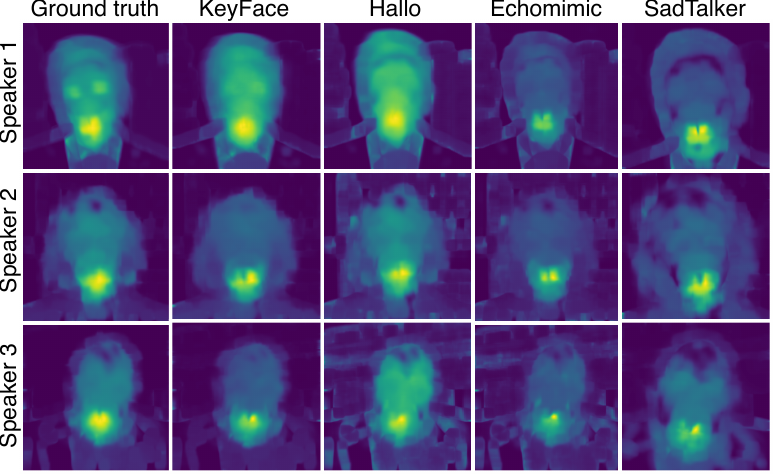}
  \caption{We show the average optical flow magnitude accross
different speakers and models.}
  \label{fig:motion_comp}
\end{figure}

\textbf{Motion.} We compare the motion generated by KeyFace to that of other existing models by analysing the average optical flow magnitude in the predicted videos in Fig.~\ref{fig:motion_comp}. AniPortrait and V-Express are excluded from this analysis, as they are conditioned on the ground-truth motion and therefore are not suitable for a fair comparison. We see that models like Hallo and EchoMimic, which rely on ReferenceNet, tend to produce background inconsistencies over time, as shown by the noisy patterns surrounding the speaker's silhouette, while SadTalker generates relatively static videos of lower quality, as indicated by a sparser optical flow map. In contrast, we find that KeyFace generates motion patterns that more closely align with those observed in real videos, outperforming other methods.

\begin{figure}[ht]
  \centering
  \includegraphics[width=\linewidth]{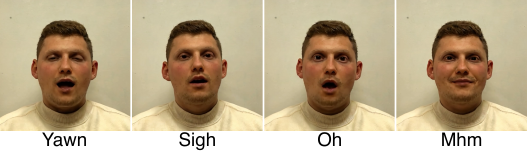}
  \caption{Examples of different NSVs generated using KeyFace, highlighting the model’s capability to handle non-speech audio.}
  \label{fig:nsv_show}
\end{figure}

\begin{figure}[ht]
  \centering
  \includegraphics[width=\linewidth]{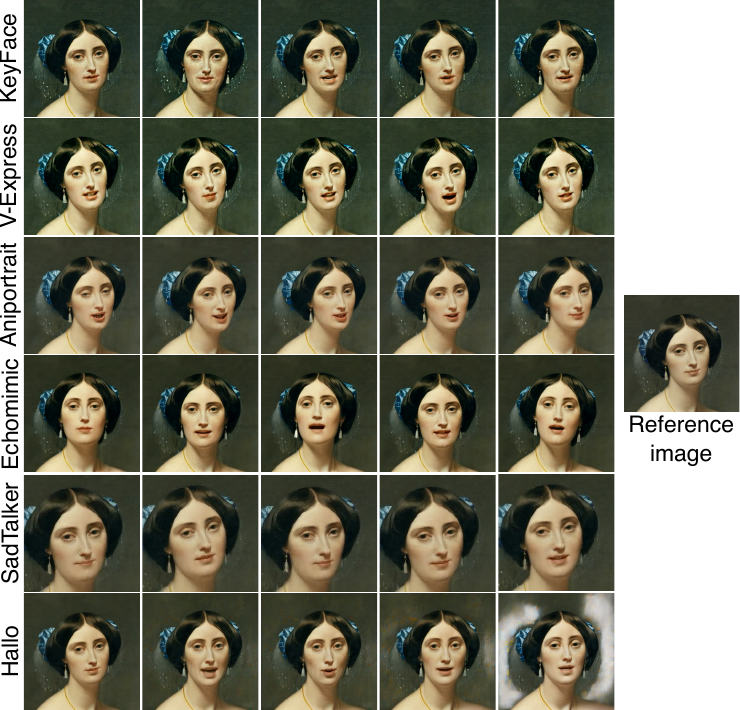}
  \caption{Results on out-of-distribution id using the same audio. Please refer to our project page for additional
video comparisons.}
  \label{fig:style_show}
\end{figure}

\textbf{Visual Quality.} Figure~\ref{fig:style_show} compares our model with other methods on the same audio input using an out-of-distribution identity frame, revealing key limitations in existing approaches. AniPortrait and SadTalker exhibit repetitive movements, V-Express treats hair accessories as background, causing unnatural head movements around the accessory, and EchoMimic introduces inconsistent head movements and background artifacts across frames. Hallo, on the other hand, produces natural motion, but suffers from error accumulation. Finally, KeyFace produces natural and varied head motion while achieving the best lip synchronization, on par with V-Express. We highlight our model’s ability to accurately animate non-speech vocalizations in Figure~\ref{fig:nsv_show}, emphasizing our holistic approach to facial animation compared to existing methods that can only handle speech. For a more comprehensive evaluation, we strongly encourage readers to refer to the supplementary material.

\section{Conclusion}
We introduce \textbf{KeyFace}, a two-stage diffusion-based framework for generating long-duration, coherent, and natural audio-driven facial animations. By leveraging an extended temporal context through keyframe generation and interpolation, our method effectively preserves temporal coherence and realism across long sequences. We further increase the expressiveness of facial animations by conditioning on continuous emotions for long-term emotional control, and adding NSVs to our training set. Experimental results demonstrate that KeyFace outperforms state-of-the-art methods across a comprehensive set of objective metrics. 
Finally, we consolidate our findings via a series of qualitative evaluations and prove that KeyFace successfully addresses key challenges such as repetitive movements and error accumulation, setting a new standard for natural and expressive animations over long durations.

{
    \small
    \bibliographystyle{ieeenat_fullname}
    \bibliography{main}
}

\clearpage
\setcounter{page}{1}
\maketitlesupplementary
\appendix


\section{Model Details}

\paragraph{Implementation}
In our experiments, both the keyframe generator and the interpolation model produce sequences of 14 frames. The keyframes are spaced by $S = 12$ frames, and the interpolation model uses two frames as conditioning. Consequently, the total number of new frames generated through interpolation is $S$. This configuration captures extended temporal dependencies while maintaining computational efficiency.

We initialize the weights of the U-Net and VAE from SVD~\cite{blattmann2023stablevideodiffusionscaling} and conduct all experiments on NVIDIA A100 GPUs with a batch size of 32 for both models. The keyframe generator is trained for 60,\,000 steps, while the interpolation model requires 120,\,000 steps due to its greater deviation from the pre-trained SVD. We use the AdamW optimizer~\cite{loshchilov2019decoupledweightdecayregularization} with a constant learning rate of $1 \times 10^{-5}$, following a 1,\,000-step linear warm-up. For inference, we use 10 steps, consistent with \cite{blattmann2023videoldm}. During training, the identity frame is randomly selected from each video clip.

Audio is sampled at 16,\,000 Hz to align with the pre-trained encoders (WavLM~\cite{wavlm} and BEATs~\cite{chen2022beatsaudiopretrainingacoustic}), while video frames are extracted at 25 fps and resized to $512 \times 512$ pixels. During training, the audio condition is randomly dropped 20\,\% of the time, and the identity condition is dropped 10\,\% of the time to strengthen the guidance effect.

We train the reduced model for autoguidance~\cite{karras2024guidingdiffusionmodelbad} with 16$\times$ fewer training steps. The default settings are summarized in Table~\ref{tab:model_params}.

\begin{table}[ht]
\centering
\resizebox{\columnwidth}{!}{
\begin{tabular}{lc}
\toprule
Parameter                & Value \\ \midrule
Keyframe sequence length ($T$)       & 14       \\
Keyframe spacing ($S$)               & 12       \\
Interpolation sequence length ($S$)  & 12       \\
Keyframe training steps              & 60,\,000   \\
Interpolation training steps         & 120,\,000 \\
Training batch size                  & 32             \\
Optimizer                            & AdamW          \\
Learning rate                        & $1 \times 10^{-5}$ \\
Warm-up steps                        & 1,\,000          \\
Inference steps                      & 10       \\
GPU used                             & NVIDIA A100    \\
Autoguidance~\cite{karras2024guidingdiffusionmodelbad} model training steps               & 120,\,000\,$/$\,16\,$=$\,7,\,500 \\ 
Audio condition drop rate for CFG~\cite{ho2022classifierfreediffusionguidance}            & 20\,\%                \\
Identity condition drop rate for CFG~\cite{ho2022classifierfreediffusionguidance}        & 10\,\%                 \\
Audio CFG~\cite{ho2022classifierfreediffusionguidance} scale & 3 \\
ID CFG~\cite{ho2022classifierfreediffusionguidance} scale & 2.33 \\
\bottomrule
\end{tabular}}
\caption{\textbf{Default model parameters and training configurations.}}
\label{tab:model_params}
\end{table}

\paragraph{Inference speed}

One limitation of our model is that it does not yet support real-time generation. Nevertheless, our two-stage approach is faster than competing diffusion-based models, particularly because it allows batching, unlike autoregressive methods. We present an inference speed comparison (Table~\ref{tab:speed}), measured in seconds per frame. Real-time inference could potentially be achieved through distillation methods (e.g., UFOGen), which we leave for future work.
\begin{table}[ht]
\centering
\resizebox{\columnwidth}{!}{%
\begin{tabular}{ccccc}
\toprule
V-Express~\cite{wang2024V-Express}            & Hallo~\cite{xu2024hallohierarchicalaudiodrivenvisual} & AniPortrait~\cite{wei2024aniportraitaudiodrivensynthesisphotorealistic} &  EchoMimic~\cite{chen2024echomimiclifelikeaudiodrivenportrait} & Keyface \\ \midrule
 3.36 &  1.9  &  0.44   & 0.76 & \textbf{0.26}   \\
\bottomrule
\end{tabular}}
\caption{Seconds per frame comparison for baseline models.}
\label{tab:speed}
\end{table}

\section{Comparison with SVD} \label{supp:svd}

Our method builds upon Stable Video Diffusion (SVD)~\cite{blattmann2023stablevideodiffusionscaling} by introducing carefully designed architectural and task-specific adaptations. These modifications distinctly set our approach apart from prior work. We highlight the primary differences below.

\paragraph{Audio Conditioning}

While SVD primarily conditions on the initial frame to predict subsequent video frames, our method extends this capability by conditioning on both an identity frame and audio inputs to drive video generation. To the best of our knowledge, we are the first to employ conditioning based on outputs from two distinct audio encoders (WavLM~\cite{wavlm} and BEATs~\cite{chen2022beatsaudiopretrainingacoustic}, allowing simultaneous processing of speech and non-speech audio.

\paragraph{Emotional Conditioning}

Unlike the original SVD architecture, our approach incorporates additional control over emotional expression. We demonstrate that training emotional models exclusively with pseudo-labels for valence and arousal achieves robust and consistent performance.

\paragraph{Loss Functions}

SVD employs only the EDM loss~\cite{karras2022elucidatingdesignspacediffusionbased}. In contrast, we use two additional pixel-space losses along with a weighted loss that specifically targets the lower region of generated images.

\paragraph{Guidance}

Whereas SVD solely employs vanilla classifier-free guidance (CFG)~\cite{ho2022classifierfreediffusionguidance}, we provide an in-depth investigation into optimal guidance techniques tailored specifically to each stage of our pipeline. We found that, for the keyframe model, assigning different CFG weights to identity and audio conditions leads to better performance and improved robustness compared to classical CFG. Additionally, since interpolation requires greater flexibility in head movement, we employed autoguidance~\cite{karras2024guidingdiffusionmodelbad} to dynamically balance guidance, resulting in enhanced overall video quality.




\section{Datasets}

\subsection{Data details} \label{sec:data_details}

Table~\ref{tab:dataset} provides an overview of the datasets used in this paper, detailing the number of speakers, videos, average video duration, and total duration for each dataset. We use a combination of publicly available datasets (HDTF~\cite{hdtf}, CelebV-HQ~\cite{zhu2022celebvhq}, CelebV-Text~\cite{yu2022celebvtext}) and our own collected data. As stated in the main paper, we use only HDTF and the collected data for training our final model. Additionally, we utilize reference frames from FEED~\cite{drobyshev2024emoportraitsemotionenhancedmultimodaloneshot} for some qualitative results.

\begin{table}[ht]
\centering
\resizebox{\columnwidth}{!}{
\begin{tabular}{lrrrr}
\toprule
Dataset         & \# Speakers & \# Videos & \multicolumn{2}{c}{Duration} \\ \cmidrule(lr){4-5}
                &             &           & Avg. (sec.) & Total (hrs.)  \\ \midrule
HDTF~\cite{hdtf}       &   264       &   318    &    139.08      &       12     \\
CelebV-HQ~\cite{zhu2022celebvhq} &   3,\,668    &  12,\,000  &      4.00       &        13    \\
CelebV-Text~\cite{yu2022celebvtext} & 9,\,109    &   75,\,307 &        6.38      &       130    \\
Collected data &   824        &   4,\,677 &      123.15      &      160     \\
Collected data (NSV) &   639        &   5,\,701 &      18.94       &       30     \\
\bottomrule
\end{tabular}
}
\caption{Overview of the datasets used in the study.}
\label{tab:dataset}
\end{table}

\subsection{Preprocessing details}

Even during our experimentation with alternative data sources in the data ablation study, we aim to obtain the highest-quality data possible. To achieve this, we propose a data preprocessing pipeline with the following steps:

\begin{itemize}
\item Extract 25 fps video and 16 kHz mono audio.
\item Discard low-quality videos based on a quality score computed using HyperIQA \cite{Su_2020_CVPR_hyperiqa}.
\item Detect and separate scenes using \href{https://github.com/Breakthrough/PySceneDetect}{PySceneDetect}.
\item Remove clips without active speakers using Light-ASD \cite{Liao_2023_CVPR_lightSDA}.
\item Estimate landmarks and poses using \href{https://github.com/1adrianb/face-alignment}{face-alignment}.
\item Crop the video around the facial region across all frames.
\end{itemize}
Using this pipeline, we curate CelebV-HQ~\cite{zhu2022celebvhq} and CelebV-Text~\cite{yu2022celebvtext}.

\begin{figure}[ht]
\centering
\includegraphics[width=\linewidth]{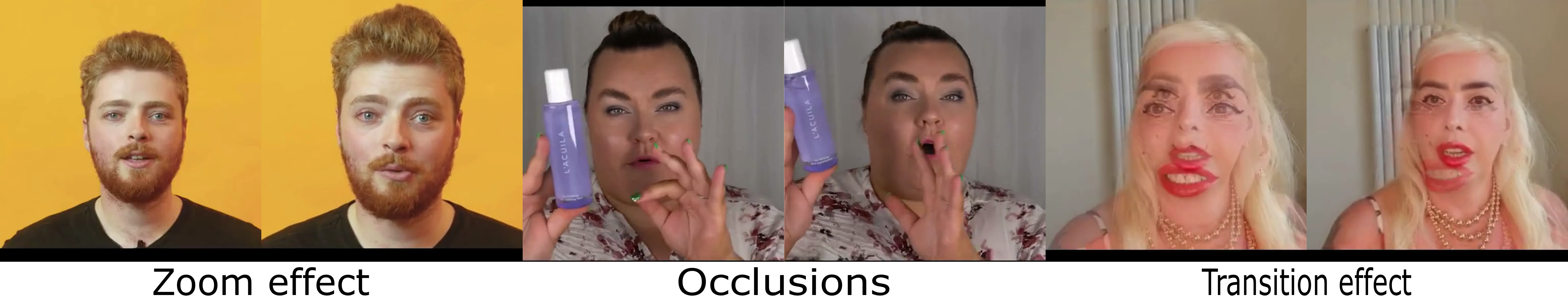}
\caption{Illustration of bad examples in CelebV-HQ~\cite{zhu2022celebvhq} and CelebV-Text~\cite{yu2022celebvtext}.}
\label{fig:bad_sampleds}
\end{figure}

However, even after filtering the datasets, we found that many samples contain editing effects and/or occlusions that are not detected. Examples include visible hands, camera movement, editing effects, and occlusions, which we found occur in 20\,\% of videos even after our cleaning process, as illustred in Figure~\ref{fig:bad_sampleds}. Since these artefacts don't correlate with speech, they can't be replicated by the model, hindering performance as shown in Section~\ref{sec:ablations}.

\section{Evaluation metrics} \label{supp:metrics}

\subsection{LipScore}

To evaluate the effectiveness of our proposed LipScore metric compared to the traditional SyncNet metric, we conduct experiments introducing controlled temporal and spatial perturbations to synchronized audio-visual data. The goal is to observe how each metric responds to these perturbations and determine which better correlates with the expected degradation in lip synchronization quality.

\paragraph{Temporal misalignment sensitivity}

In the first set of experiments, we introduce temporal misalignments by shifting the ground truth video temporally. The time shifts range from 0 milliseconds (ms) to 1000 ms.

\begin{figure}[ht]
\centering
\includegraphics[width=\linewidth]{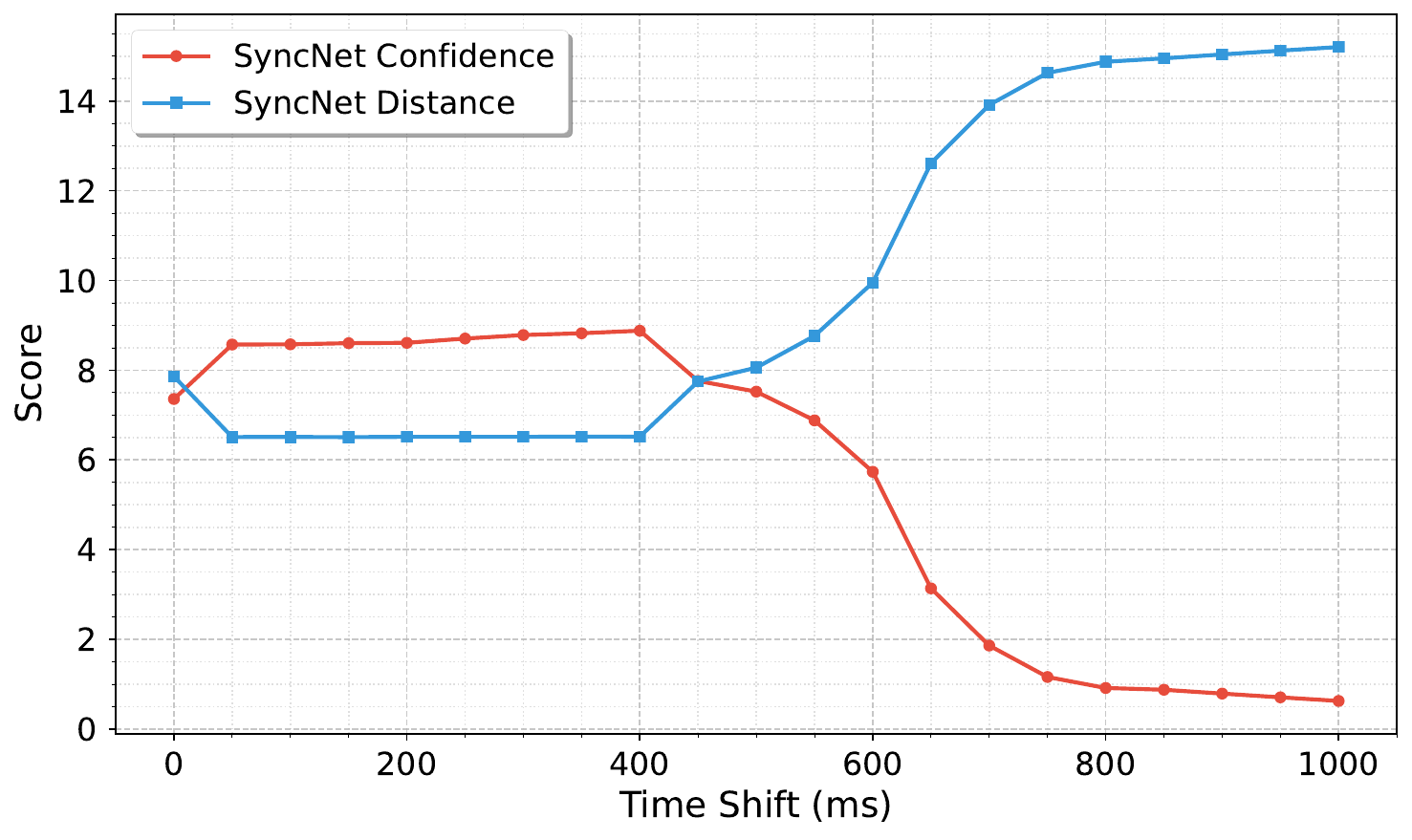}
\caption{SyncNet Confidence and SyncNet Distance as functions of time shift (ms).}
\label{fig:timeshift_syncnet}
\end{figure}

Figure~\ref{fig:timeshift_syncnet} illustrates the behavior of SyncNet Confidence and SyncNet Distance as functions of the time shift. We observe that SyncNet Confidence and Distance remain constant up to approximately 400 ms and only start to change significantly beyond this point. This behavior is undesirable, as even small misalignments (e.g., 100–200 ms) should result in a noticeable decrease in confidence and an increase in distance.

\begin{figure}[ht]
\centering
\includegraphics[width=\linewidth]{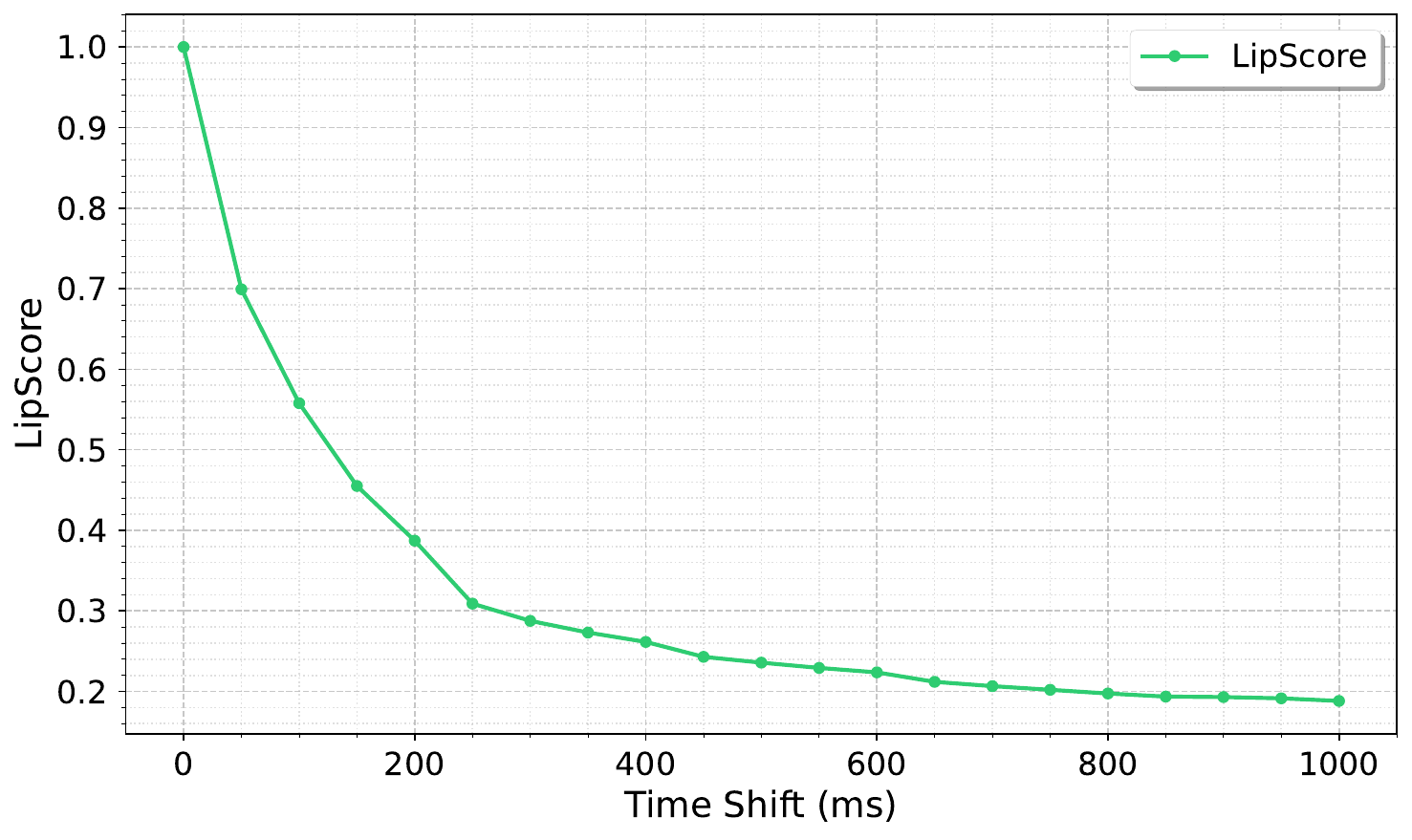}
\caption{LipScore as a function of time shift (ms).}
\label{fig:timeshift_lipscore}
\end{figure}

In contrast, Figure~\ref{fig:timeshift_lipscore} shows the LipScore metric’s response to the same range of time shifts. LipScore exhibits a stable and consistent decrease in score as the time shift increases. It begins to penalize even small temporal perturbations, with a sharp decline at smaller offsets, and stabilizes at lower scores as larger misalignments are introduced.
This behavior aligns with the expected characteristics of a robust lip synchronization metric, demonstrating continuous sensitivity to temporal misalignments without erratic or overly abrupt changes.

\paragraph{Robustness to spatial perturbations}

We evaluate the robustness of the metrics to spatial transformations by introducing horizontal shifts and rotations to the video frames.

\begin{figure}[ht]
\centering
\includegraphics[width=\linewidth]{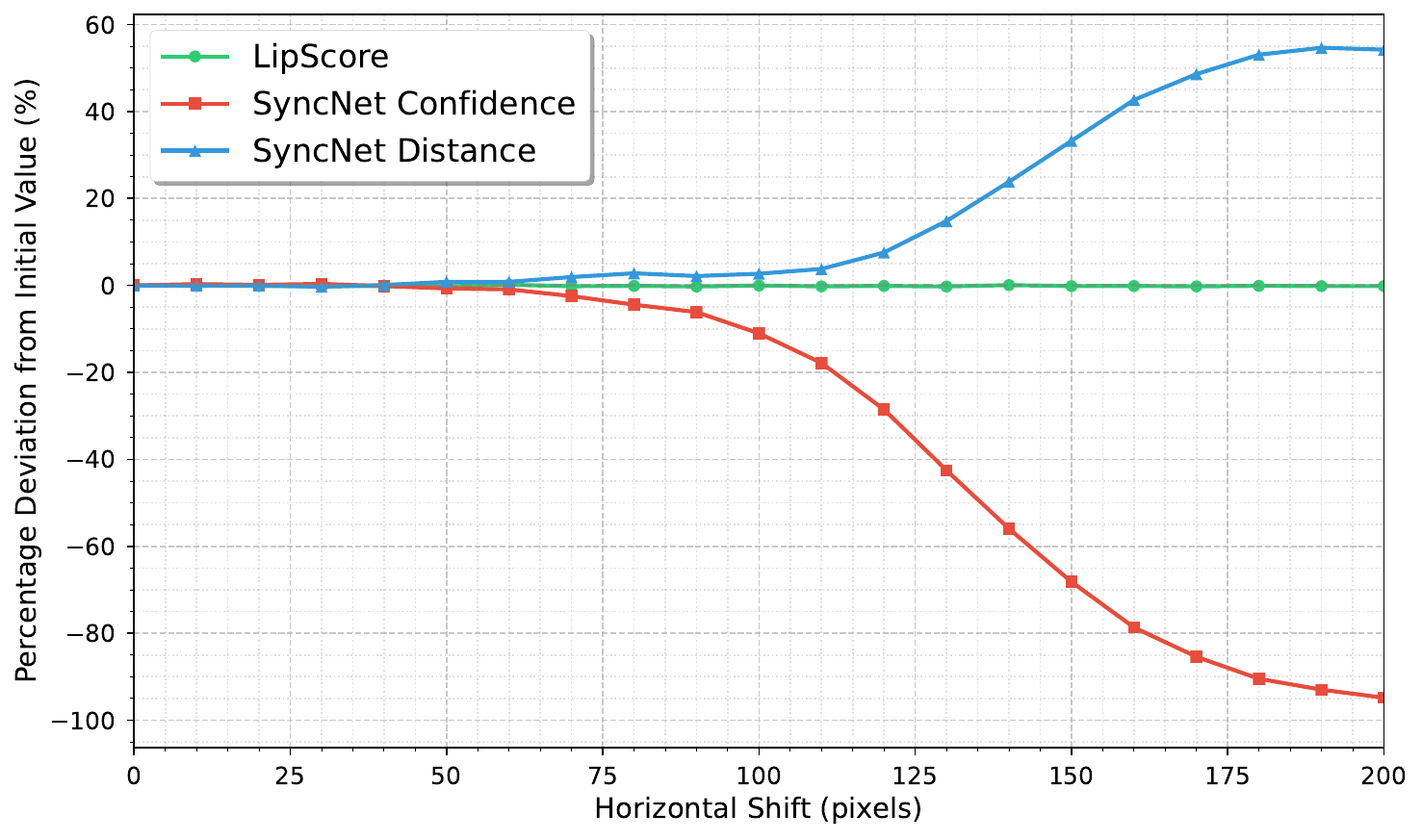}
\caption{Effect of horizontal shifts on LipScore, SyncNet Confidence, and SyncNet Distance. The plot shows the percentage deviation from the initial value as the horizontal shift increases.}
\label{fig:hor_shift}
\end{figure}

Figure~\ref{fig:hor_shift} illustrates the percentage deviation from the initial metric values as horizontal shifts increase. LipScore remains stable, exhibiting minimal deviation across the range of horizontal shifts, indicating its robustness to this type of spatial perturbation. In contrast, SyncNet Confidence and SyncNet Distance show significant deviations starting at a shift of 75 pixels, highlighting their sensitivity to horizontal displacements.

\begin{figure}[ht]
\centering
\includegraphics[width=\linewidth]{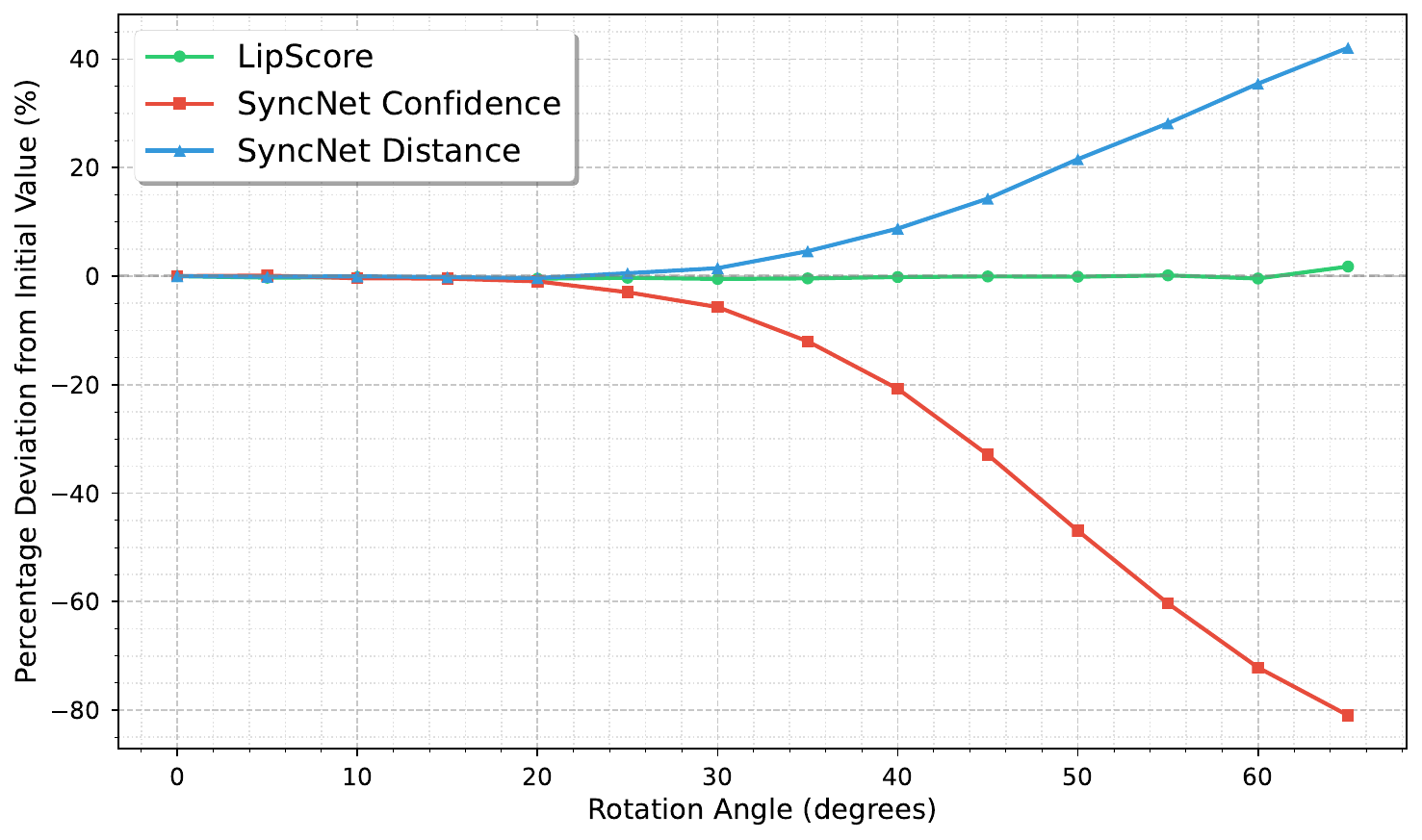}
\caption{Effect of rotation angles on LipScore, SyncNet Confidence, and SyncNet Distance. The plot shows the percentage deviation from the initial value as the rotation angle increases.}
\label{fig:rot_shift}
\end{figure}

Similarly, Figure~\ref{fig:rot_shift} shows the percentage deviation in metric values as the rotation angle of the video frames increases. LipScore again demonstrates robustness, with negligible changes in its values even as the rotation angle grows. In contrast, SyncNet Confidence and SyncNet Distance exhibit substantial deviations starting at 20 degrees, indicating that these metrics are more adversely affected by rotational transformations.

\paragraph{WER on unseen datasets}

We additionally evaluate our state-of-the-art lipreader~\cite{Ma_2023} on HDTF and find that it achieves a 21\,\% WER, demonstrating strong performance on unseen data and further supporting LipScore’s validity.

\subsection{Non-speech vocalization classifier}

We introduce the Non-Speech Vocalization (NSV) Classifier as part of our evaluation methodology. This not only highlights the limitations of pre-trained speech-driven animation methods but also demonstrates the capabilities of our model in generating realistic NSV sequences. The model processes video inputs and classifies them into one of eight NSV types, plus speech.

\paragraph{Architecture}

\begin{figure}[ht]
\centering
\includegraphics[width=\linewidth]{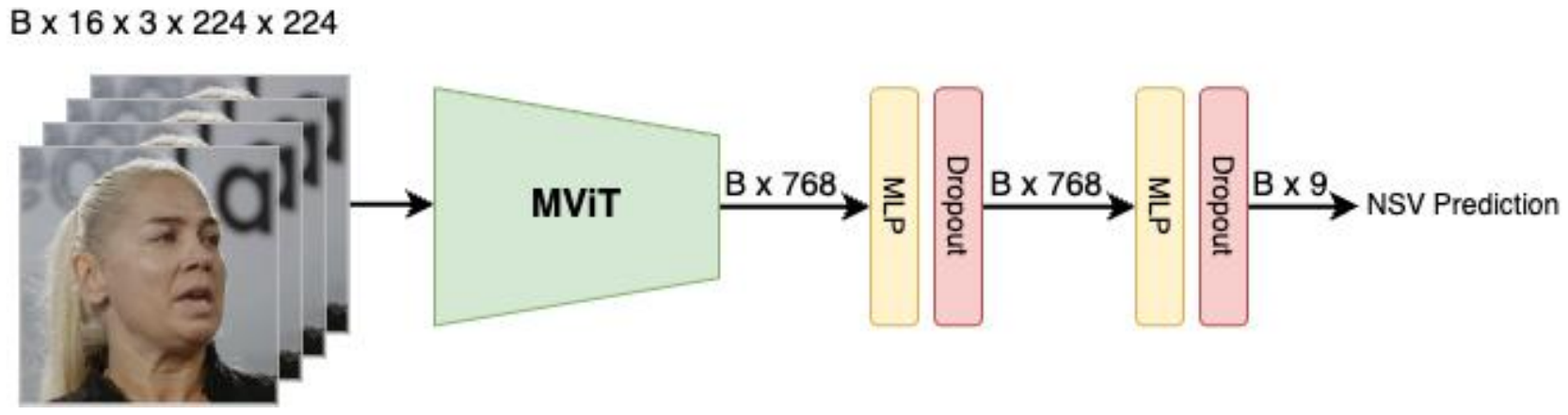}
\caption{The architecture used for the Non-Speech Vocalization Classifier. The batch size is denoted as B.}
\label{fig:nsv_classifier_arch}
\end{figure}

The architecture of the system is presented in Fig.~\ref{fig:nsv_classifier_arch}. We employ a Multiscale Vision Transformer (MViTv2)~\cite{li2022mvitv2improvedmultiscalevision} backbone, augmented with two linear layers and a dropout layer with a dropout probability set to 0.2. The MViTv2 model, pre-trained on the Kinetics dataset~\cite{kay2017kineticshumanactionvideo}, achieves a top-5 accuracy of 94.7\,\%.

\paragraph{Training}

Our model is trained using a dataset containing video clips of eight different NSV types and speech. The eight NSV classes are: \textit{"Mhm"}, \textit{"Oh"}, \textit{"Ah"}, \textit{coughs}, \textit{sighs}, \textit{yawns}, \textit{throat} \textit{clears}, and \textit{laughter}. During the training process, video clips corresponding to any of these classes are fed into the model. We train using the AdamW optimizer with a learning rate of $1\times10^{-4}$, $\beta_1=0.9$, and $\beta_2=0.999$. The cross-entropy loss is employed as the loss function.

Our model achieves an F1 score of 0.7 across these nine classes, demonstrating its effectiveness in classifying various NSVs and speech.

\paragraph{NSVs performance boundaries}

\begin{figure}[ht]
  \centering
  \includegraphics[width=\linewidth]{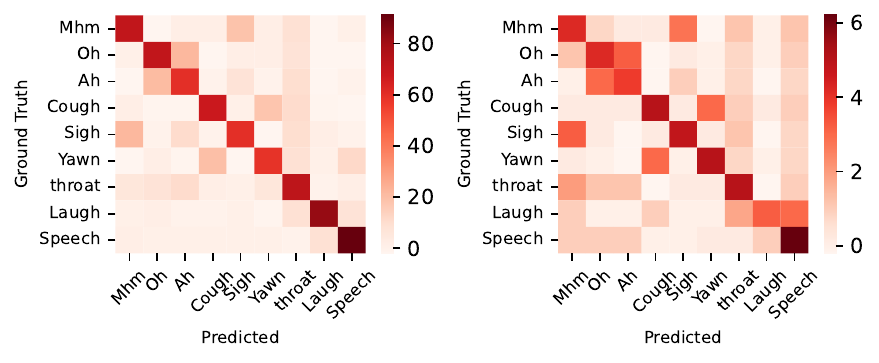}
  \caption{NSV confusion matrix for generated (left) and validation (right) videos.}
  \label{fig:nsv_confusion} 
\end{figure}

To demonstrate and understand the effectiveness of  $NSV_{acc}$ across individual NSVs, we present a confusion matrix on the validation set of the data used to train $NSV_{acc}$ (Fig.\ref{fig:nsv_confusion}, left). Although the model achieves good overall performance, certain NSVs are frequently confused, such as “Oh” with “Ah,” “Sigh” with “Mhm,” and “Yawn” with “Cough.” 

Additionally, we demonstrate that our model can generate visually distinct NSVs (Fig.\ref{fig:nsv_confusion}, right) with few confusions by generating 10 videos per NSV category and speech.

\section{User study details} \label{supp:user}

To evaluate the performance of our proposed method, KeyFace, against existing baselines, we conduct a comprehensive user study. Participants view pairs of talking face videos and select the one they find more realistic. This section summarizes the results of the pairwise comparisons and the derived metrics.

\begin{figure}[ht]
  \centering
  \includegraphics[width=\linewidth]{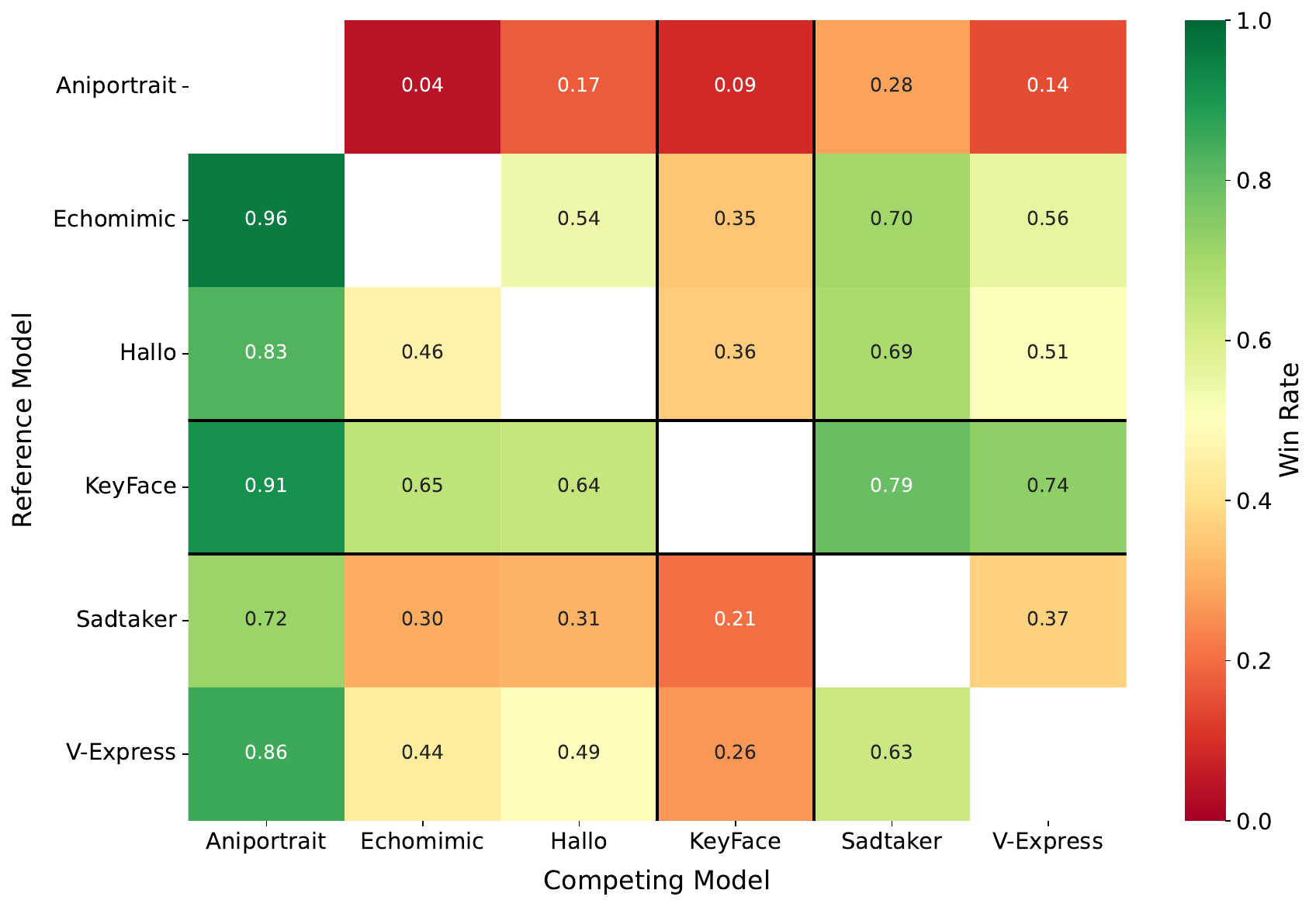}
  \caption{Pairwise win rates between reference (rows) and competing models (columns). Green indicates higher, Red lower win rates.}
  \label{fig:winrate_mat}
\end{figure}

\paragraph{Pairwise Win Rates:}
The pairwise win rate matrix is presented in Figure~\ref{fig:winrate_mat}. Each cell represents the proportion of times the reference model (rows) is preferred over the competing model (columns). Green indicates a high win rate for the reference model, while red represents a lower win rate. KeyFace is consistently preferred over baseline models, achieving a win rate of at least 64\,\% against all other methods.

\begin{figure}[ht]
  \centering
  \includegraphics[width=\linewidth]{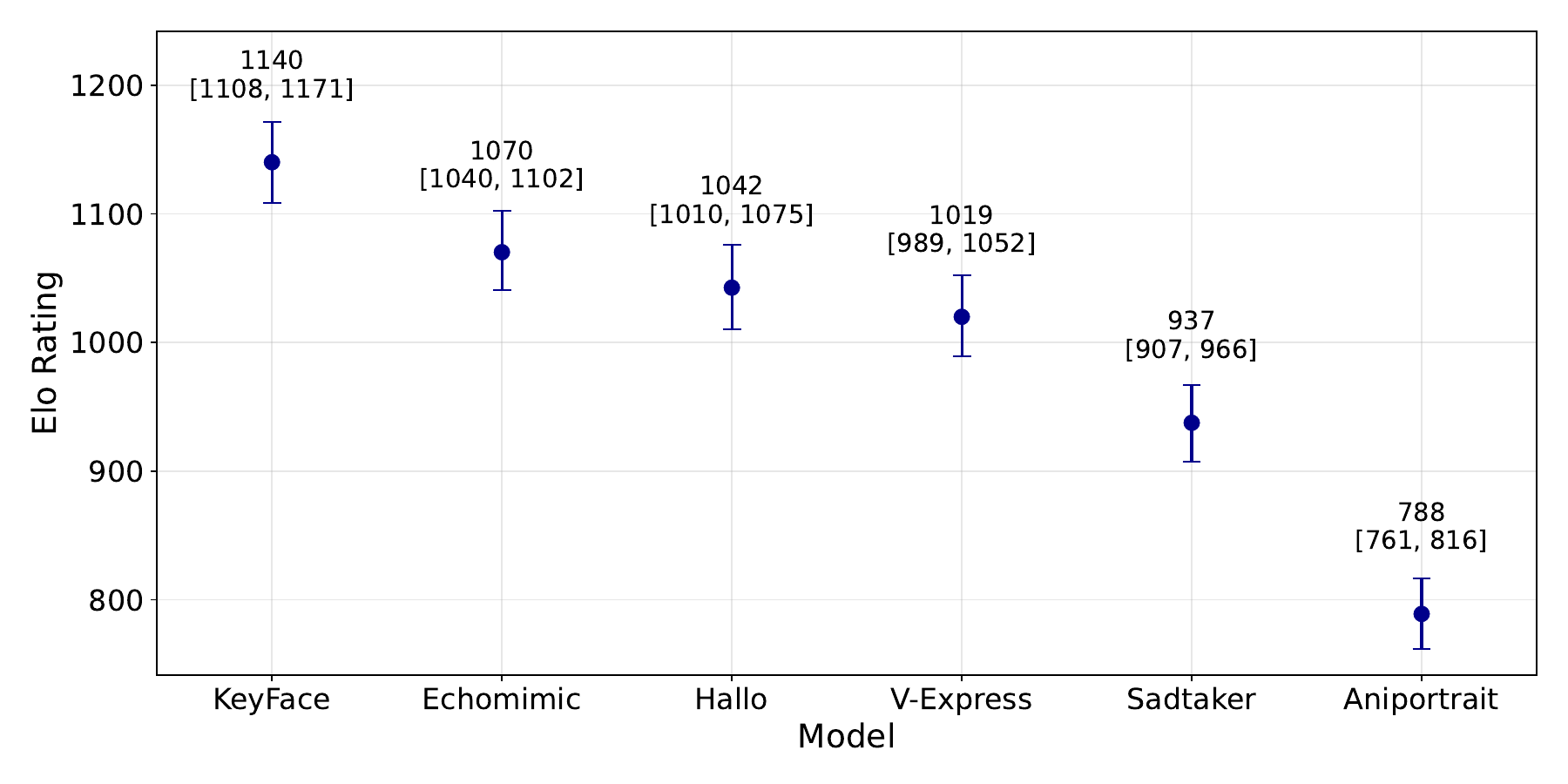}
  \caption{\textbf{Elo ratings for all models} with 95\,\% confidence intervals. Higher ratings indicate better overall performance.}
  \label{fig:err_bar}
\end{figure}

\paragraph{Elo ratings:}
Figure~\ref{fig:err_bar} presents the Elo ratings for all models with 95\,\% confidence intervals. KeyFace achieves the highest Elo rating, significantly outperforming the baselines, demonstrating its effectiveness in generating high-quality talking face animations.

\begin{figure}[ht]
  \centering
  \includegraphics[width=\linewidth]{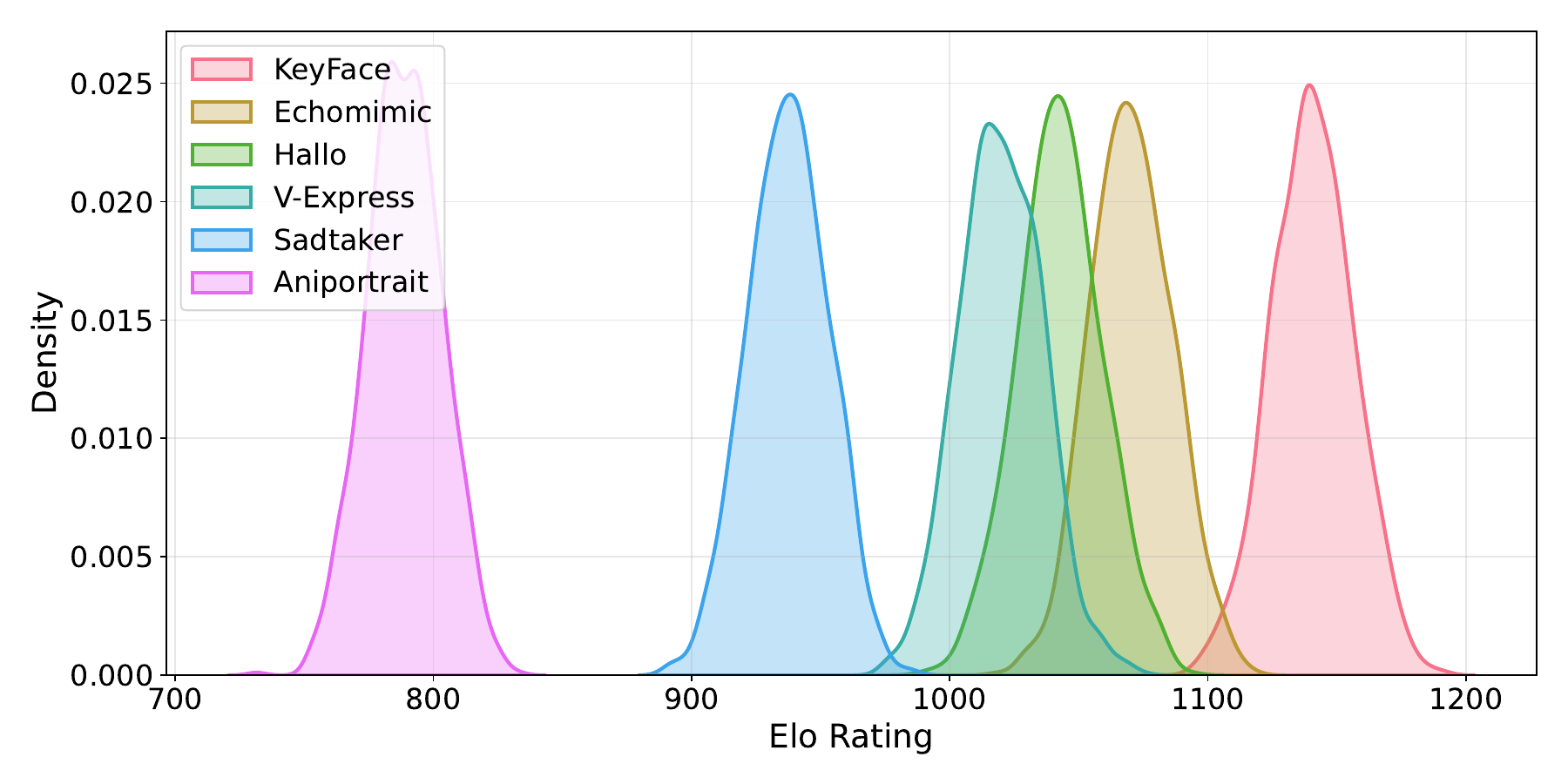}
  \caption{\textbf{Density distributions of Elo ratings for all models.} Peaks indicate the most probable performance levels, with higher ratings reflecting better performance.}
  \label{fig:err_dist}
\end{figure}

\paragraph{Elo rating distributions:}
The density distributions of Elo ratings are shown in Figure~\ref{fig:err_dist}. KeyFace exhibits a sharp, high-density peak at the upper end, highlighting its robustness and consistent user preference across evaluation scenarios. Echomimic, V-Express, and Hallo show significant overlap in their results, while Aniportrait and SadTalker consistently receive lower ratings.

\section{Additional ablation}

\paragraph{Audio mechanisms}

\begin{table}[ht]
\centering
\resizebox{\columnwidth}{!}{%
\begin{tabular}{lccc}
\toprule
Method             & FID $\downarrow$ & FVD $\downarrow$ & LipScore $\uparrow$ \\ \midrule
w/o cross attention &           16.95                 & 167.39               & 0.35                  \\
w/o timestep        &           17.20                &  176.83               & 0.28                  \\ 
\rowcolor{Gray!40} cross attention + timestep & \textbf{16.76}   & \textbf{137.25} & \textbf{0.36} \\ 
\bottomrule
\end{tabular}}
\caption{\textbf{Audio conditioning ablation} on HDTF~\cite{hdtf}: “Cross attention” refers to incorporating audio through a cross-attention mechanism, while “timestep” refers to adding the audio embeddings to the timestep embeddings. The best results are highlighted in \textbf{bold} and default settings are
highlighted in \colorbox{Gray!40}{gray} on all tables.}
\label{tab:audio_mech}
\end{table}

Table~\ref{tab:audio_mech} presents an ablation study on the impact of different audio conditioning mechanisms on video generation quality. The results show that the audio timestep plays a critical role in achieving accurate lip synchronization, as removing it (row “w/o timestep”) results in the lowest LipScore and the highest FVD. Adding cross attention alone improves video quality but only marginally enhances the LipScore compared to when the timestep is absent. The best performance is achieved when both cross attention and audio timestep embeddings are used together, leading to the lowest FID, significantly lower FVD, and the highest LipScore. This indicates that while audio timestep embeddings are essential for achieving good lip synchronization, the addition of cross attention further enhances the overall quality of the generated videos by improving visual coherence and temporal consistency.

\paragraph{Training on HDTF only} \label{supp:only_hdtf}

To ensure a fair comparison with baseline models, we retrain our model exclusively on publicly available data (i.e. HDTF~\cite{hdtf}), removing all non-public sources. Although this leads to a decrease in performance, our model still outperforms baseline methods trained on larger datasets. We emphasize that most existing methods rely on private datasets; therefore, to maintain fairness, we curated our dataset to have comparable scale in terms of total hours and number of speakers as described in Section~\ref{sec:data_details}.

\begin{table}[ht]
\centering
\begin{tabular}{lccc}
\toprule
Method             & FID $\downarrow$& FVD $\downarrow$& LipScore $\uparrow$ \\ \midrule
KeyFace (HDTF only) &   19.49  &  165.06   &  0.28    \\
\bottomrule
\end{tabular}
\caption{Results of pipeline trained on HDTF only.}
\label{tab:only_htf}
\end{table}

\section{Limitations}

\begin{figure}[ht]
\centering
\includegraphics[width=\linewidth]{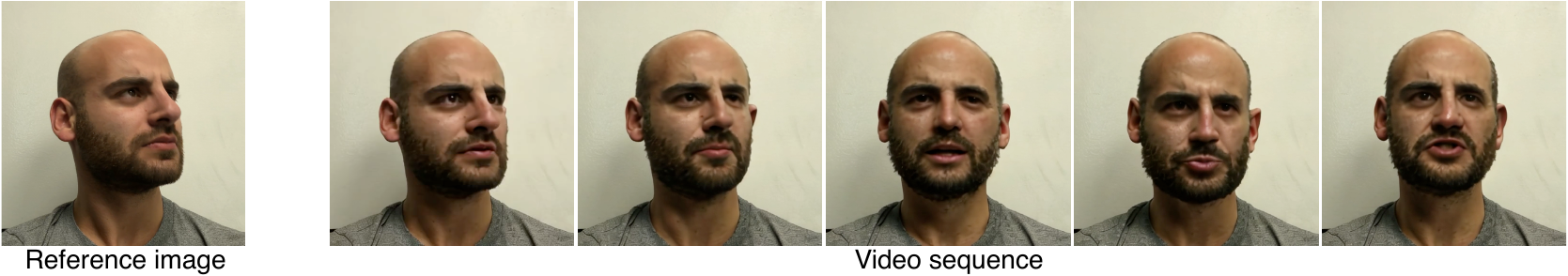}
\caption{An example showcasing KeyFace's limitations in handling extreme head poses.}
\label{fig:limitation}
\end{figure}

One key limitation of our model, which it shares with all baseline methods, is its performance when the initial frame exhibits an extreme head pose. This issue primarily stems from the lack of training data containing such extreme poses, resulting in difficulties in reconstructing the occluded or unseen parts of the face. As illustrated in Figure~\ref{fig:limitation}, although the model can generate plausible videos with accurate lip synchronization, it partially loses the identity of the reference image in these scenarios. Additional failure cases involving challenging reference frames are provided in the supplementary videos.

\section{Additional qualitative results}

\begin{figure}[ht]
\centering
\includegraphics[width=\linewidth]{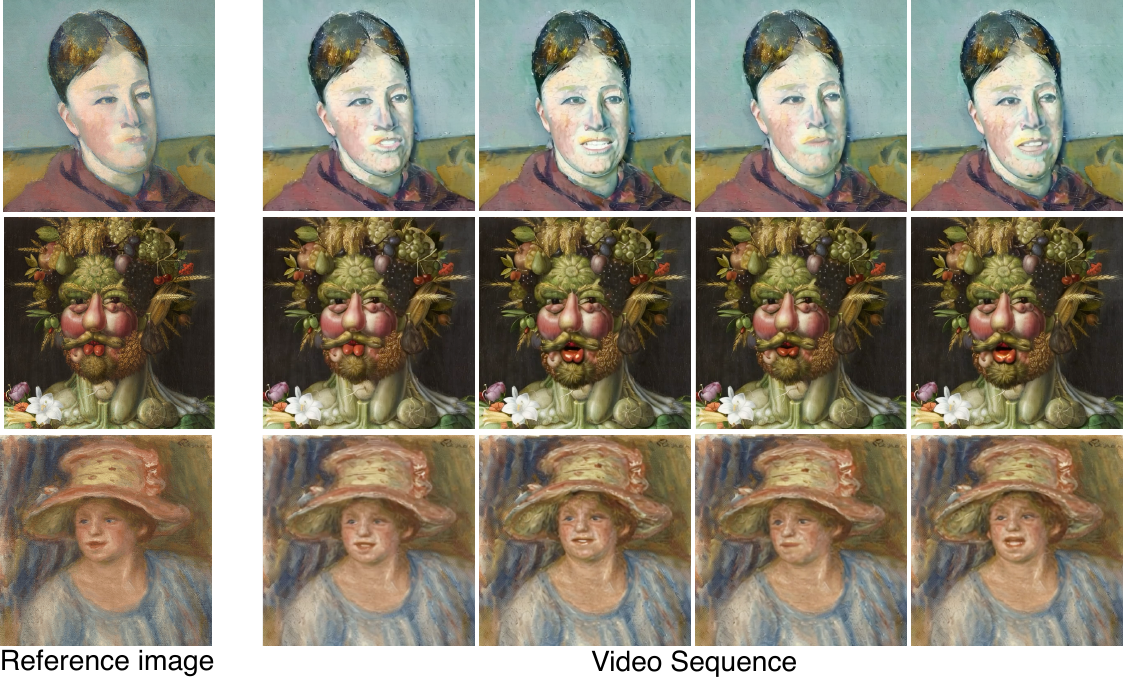}
\caption{We present a set of examples with \textbf{out-of-distribution} reference frames.}
\label{fig:odd}
\end{figure}

To further demonstrate the effectiveness of our method, we provide \textbf{example videos generated by KeyFace} (as well as competing methods, for comparison) in the supplementary material:

\begin{itemize}

\item \textbf{Non-speech vocalizations comparison.} We evaluate the model’s ability to handle eight distinct NSVs and compare its performance with baseline methods, highlighting the limitations of current state-of-the-art models and the strengths of our approach. For a fair comparison, all examples maintain a neutral emotional tone.

\item \textbf{Speech and NSV comparison.} We demonstrate the model’s capability to generate both speech and NSVs within the same video, comparing its performance to other approaches. The results showcase the holistic nature of our method, particularly in contrast to baseline models. We maintain a neutral emotional tone for consistency.

\item \textbf{Side-by-side comparison.} We present side-by-side comparisons between KeyFace and baseline models, showcasing KeyFace’s superior performance in generating realistic and expressive facial animations.
\item \textbf{Emotion interpolation.} We showcase transitions between different emotional states, emphasizing the model’s ability to capture subtle and nuanced expressions.
\item \textbf{Out-of-distribution robustness.} Figure~\ref{fig:odd} illustrates the model’s robustness in handling non-human faces, demonstrating successful generalization to a variety of input conditions.
\item \textbf{Expanded KeyFace examples.} We provide additional videos featuring KeyFace-generated animations in English and other languages, highlighting the model’s generalization capabilities across different linguistic contexts.
\end{itemize}

\end{document}